%% file: neurips_2026.tex
\documentclass{article}
\usepackage{graphicx}

\usepackage{amsmath}
\usepackage{multirow}
\usepackage[table]{xcolor}
\newcommand{\hl}[1]{\cellcolor{gray!15}#1}
\usepackage{asymptote}
\usepackage{fancyvrb}

 \usepackage[main, final]{neurips_2026}

\usepackage[utf8]{inputenc} 
\usepackage[T1]{fontenc}    
\usepackage{hyperref}       
\usepackage{url}            
\usepackage{booktabs}       
\usepackage{amsfonts}       
\usepackage{nicefrac}       
\usepackage{microtype}      
\usepackage{xcolor}         

\title{Memorize Theorems, Not Instances: Probing SFT Generalization through Mathematical Reasoning}

\author{
Ruiying Peng$^{1}$\thanks{\tt\small pry24@mails.tsinghua.edu.cn\\ \url{https://github.com/Ivine11/Theorem-SFT.git}.\\Work done during Xueyu Wu's tenure at Huawei Technologies.}
\enspace
Mengyu Yang$^{1}$
\enspace
Jing Lei$^{2}$
\enspace
Xiaohui Li$^{2\dagger}$
\enspace
Xueyu Wu$^{2}$
\enspace
Xinlei Chen$^{1\dagger}$ \\
$^{1}$Tsinghua Shenzhen International Graduate School\\
$^{2}$Huawei Technologies\\
{\small $\dagger$ Corresponding authors}
}

\usepackage{subcaption}

\begin{document}

\maketitle

\input{sec/0_abstract}
\input{sec/1_intro}

\input{sec/3_WhySftFail}

\input{sec/4_TheoremLearning}

\input{sec/5_Results}
\input{sec/2_relatedwork}
\input{sec/6_conclusion}

\begin{ack}
We thank all co-authors for their support and contributions to this work. We are especially grateful to Xiaohui Li for her invaluable help in revising and improving the manuscript. We also thank Mengyu Yang for assisting with training-code execution.

We gratefully acknowledge the computational support provided by Huawei, the Natural Science Foundation of China under Grant 62371269, and the Shenzhen Science and Technology Program under Grant ZDCY202517012.
\end{ack}

\clearpage
\bibliographystyle{unsrtnat}
\bibliography{main}






\appendix

\input{sec/7_appendix}

\clearpage
\input{checklist.tex}

\end{document}

%% file: sec/0_abstract.tex
\begin{abstract}

Supervised Fine-Tuning (SFT) is widely used for task-specific adaptation, yet recent work shows it systematically undermines reasoning generalization. We argue the root cause is not memorization itself, but its target: vanilla SFT drives models to exploit and memorize spurious surface correlations in problem-solution pairs, leaving them brittle to superficial input variations.
To address this, we propose \textbf{Theorem-SFT}, which reorients supervision toward explicit theorem application by teaching models how rules are invoked rather than what answers look like. Theorem-SFT yields consistent gains across benchmarks and model families: +8.8\% on MATH (LLaMA3.2-3B-Instruct) and +20.27\% on GeoQA (Qwen2.5-VL-7B-Instruct) without modality-specific re-training. Fine-tuning MLP layers alone matches full-layers performance, implicating feed-forward components as the primary locus of reasoning rules. Our findings reframe the debate: Generalization failures stem not from memorization as a mechanism, but from memorizing the wrong inductive targets.


\begin{figure}[htbp] 
  \centering      
  \includegraphics[width=1.0\linewidth]{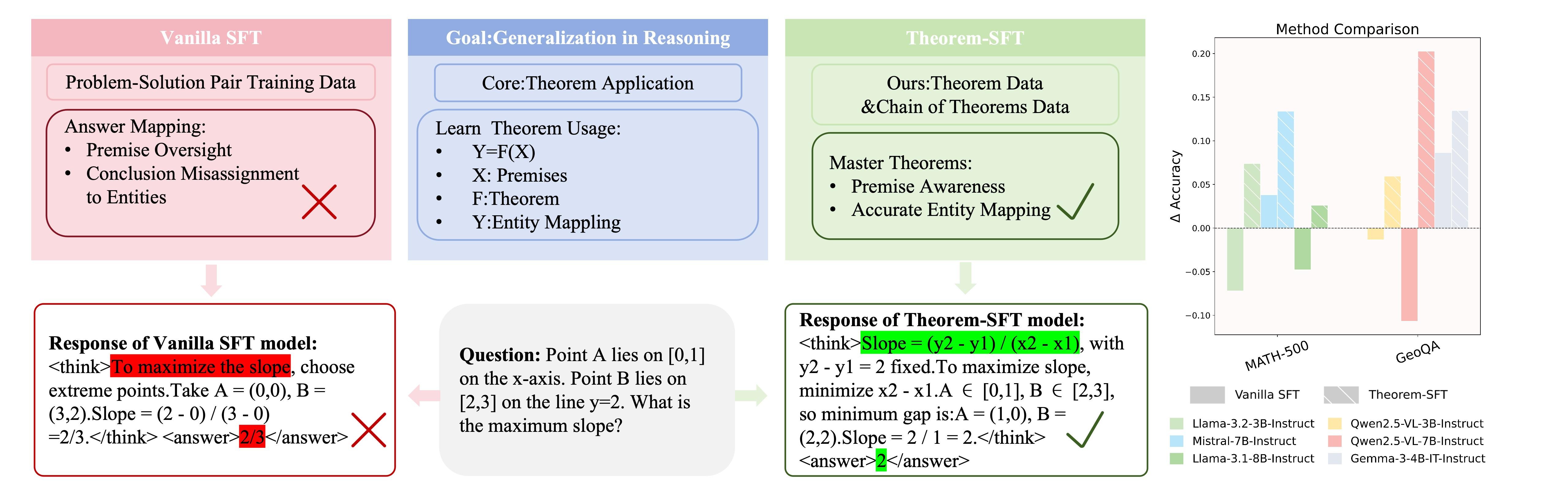} 
\caption{
Left: Comparison between Vanilla SFT and Theorem-SFT (ours), illustrating the goal of Theorem-SFT: learning theorems to improve reasoning generalization. 
Right: Changes in accuracy ($\Delta$ accuracy) relative to the baseline for models trained with Vanilla SFT and Theorem-SFT. 
The significant accuracy gains and corrected reasoning processes demonstrate that Theorem-SFT enables both LLMs and VLMs to apply theorems based on their premises, leading to more accurate and reliable solutions with strong generalization ability. The QA pairs shown in the figure are simplified abstractions derived from real examples to highlight the core issues; the original QA instances can be found in the appendix.
}
  \label{fig:fig1}
\end{figure}


\end{abstract}

%% file: sec/1_intro.tex
\section{Introduction}\label{sec:intro}


The
supervised fine-tuning (SFT) is the dominant paradigm for adapting large language models~(\cite{llmachiam2023gpt,llmbai2025qwen3,llmbrown2020language,llmjiang2023mistral7b,llmlu2025ovis2,llmsellergren2025medgemma,llmtouvron2023llama}) to specific tasks. Yet accumulating evidence~(\cite{ouyang2022training,wei2021finetuned,wang2023self}) shows that SFT systematically undermines reasoning generalization~(\cite{wei2022chain,kojima2022large,yao2023tree}): models fine-tuned on standard SFT data, typically problem-solution pairs, often achieve impressive in-distribution performance but fail catastrophically when the underlying rules must be applied under novel conditions.


Prior work~(\cite{chu2025sft,chen2025sft})  attributes these failures to memorization, arguing that SFT primarily retains problem formats, lexical co-occurrence patterns, and solution trajectories from training instances. 
We argue, however, that the issue is not whether models memorize, but what they are driven to memorize. Trained on problem-solution pairs, models learn to exploit co-occurrence patterns between problem templates and solution steps, rather than learning the preconditions and variable bindings that govern theorem application.
\begin{figure}[htbp] 
\centering 
\includegraphics[width=0.9\linewidth]{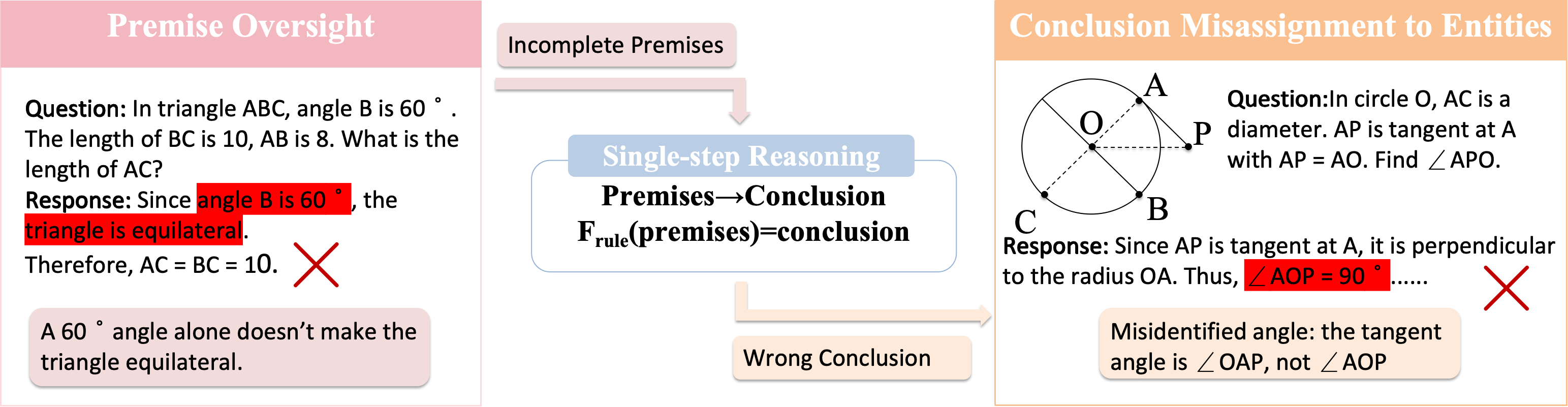} 
\caption{
Errors in single-step reasoning. We identify two main failure modes: premise oversight and entity misassignment in conclusions. The QA pairs are simplified abstractions from real examples to highlight these issues, with full instances provided in the appendix.
}
\label{fig:Error}
\end{figure}
We use mathematical reasoning as a diagnostic setting to examine this distinction. Mathematics provides a natural testbed: its theorems are explicit, enumerable, and governed by strict preconditions, making it possible to precisely characterize the difference between reproducing solution trajectories and correctly selecting and applying theorems under verified preconditions. Through systematic error analysis on both baseline and vanilla SFT models, i.e., models as is and models finetuned with mathematical reasoning QA pairs, we identify two recurring failure patterns:

\textbf{Premise Oversight}: applying a theorem without first verifying that its preconditions hold, causing incorrect results when problem conditions change even slightly, as shown in Figure~\ref{fig:Error}.

\textbf{Conclusion Misassignment to Entities}: incorrectly binding theorem variables to problem entities, a failure to recognize abstract semantic roles such as vertex, center, or tangent point, rather than a failure to recall symbol names, as shown in Figure~\ref{fig:Error}.

These patterns indicate that models fail to encode the structural components of theorems: the specific conditions that must hold and the mapping of entities to variable roles. Instead, models learn a ``shortcut'' association between keyword patterns and theorem labels, which collapses as soon as entity names are permuted or preconditions are subtly altered.

This points to a more fundamental question: what is the right inductive target for learning to reason? In human mathematical education, theorems are rarely taught solely through implicit exposure to worked examples. Instead, they are stated explicitly with named preconditions and defined variable roles. This allows learners to acquire the structure of the rule first. In this framework, applying a known theorem to a new problem is an act of deduction, whereas deriving a general rule from scattered, implicit instances, which is the task vanilla SFT forces upon models, is the much harder problem of rule induction.

Inspired by this, we propose Theorem-SFT, a training paradigm that makes the rules explicit in the training signals. Rather than training models to reproduce end-to-end solution trajectories, we teach them the structural properties of theorems: when they apply and how their variables map to entities. We construct two complementary datasets: a Theorem-Level dataset for learning applicability conditions, and a Theorem-Chain dataset for modeling multi-step logic. 

Crucially, Theorem-SFT achieves significant performance gains without direct exposure to problem instances during the theorem-learning phase, as shown in Figure~\ref{fig:fig1}. Our contributions are as follows:

\begin{itemize}
\item Mechanistic Error Analysis: We demonstrate that SFT failures specifically arise from the inability to encode theorem structure, including applicability conditions and variable roles, which manifests as premise oversight and conclusion entity misalignment.

\item Theorem-SFT Paradigm: We introduce a method for training on structured theorem representations and a corresponding mathematical theorem-learning dataset. Theorem-SFT yields a +8.8\% improvement on MATH-500 (LLaMA-3.2-3B) and a +20.27\% gain on GeoQA (Qwen2.5-VL-7B) without modality-specific re-training, showing robust generalization across model families (Figure \ref{fig:fig1}).
\item Localization of Reasoning Rules: We provide mechanistic analysis showing that restricting updates to MLP layers matches full-parameter performance. This suggests that MLP layers serve as the primary "rule library" for mathematical knowledge in transformers.
\end{itemize}

Taken together, our findings suggest that the generalization gap in SFT stems not from memorization itself, but from memorizing incorrect inductive targets, and that reorienting supervision toward explicit theorem representations provides a tractable path toward closing this gap.

%% file: sec/3_WhySftFail.tex
\section{Why does Vanilla SFT fail in reasoning?}
\label{sec:whyFail}

We identify two recurring failure modes of vanilla SFT, which it not only fails to address but often exacerbates. These issues stem from a mismatch in the inductive bias: formal reasoning depends on structured rule execution, whereas vanilla SFT primarily latches onto surface-level co-occurrence patterns. As discussed in Sec.~\ref{sec:intro}, vanilla SFT essentially solves an inverse problem, recovering general rules from implicit problem-solution pairs. This framing reduces reasoning to pattern matching over observed signals, rather than principled deduction. Consequently, the resulting models exhibit systematic failures that mirror the underlying structure of theorem application.

\begin{figure}[htbp] 
  \centering      
  \includegraphics[width=1.0\linewidth]{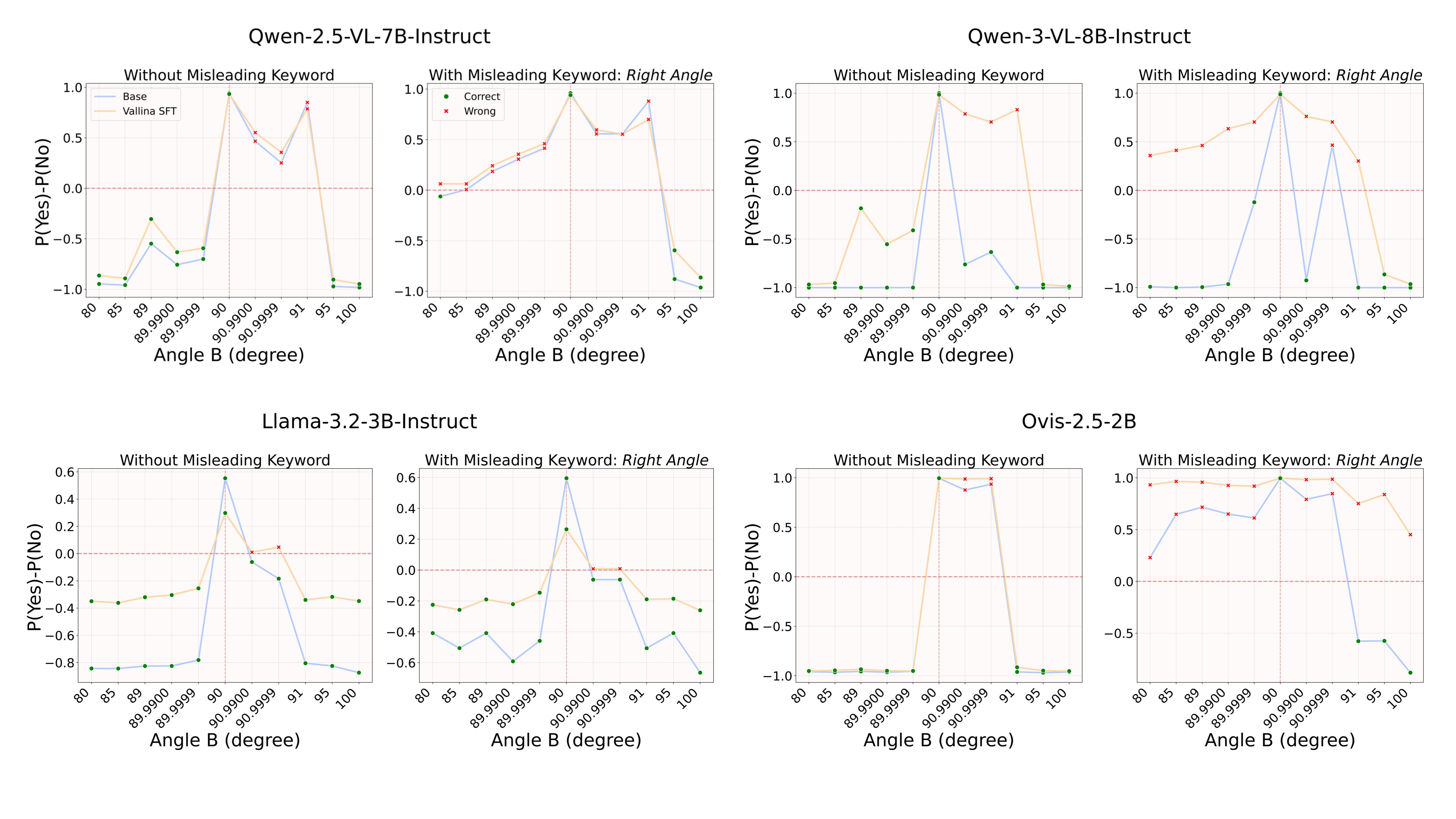} 
\caption{
Model responses to the applicability of the \textit{Pythagorean theorem} under different angle settings, illustrating premise oversight. The x-axis shows angle $B$ (in degrees), and the y-axis shows $P(\text{Yes}) - P(\text{No})$. Each pair of plots compares a standard question with a misleading variant containing the keyword ``right angle'', even when $B \neq 90^\circ$. The correct answer is ``Yes'' only when $B = 90^\circ$.
}
  \label{fig:KeywordBias}
\end{figure}

\subsection{Premise Oversight}

A theorem is only valid under its premises. However, as shown in Figure~\ref{fig:KeywordBias}, models often trigger rule application based on lexical cues without verifying whether the required conditions are satisfied. For example, the presence of ``right triangle'' frequently leads to applying the Pythagorean theorem even when the triangle does not contain a $90^\circ$ angle.

Our logit analysis ($P(\text{Yes}) - P(\text{No})$) quantifies this tendency, showing that models still favor rule application in inconsistent cases and are highly sensitive to misleading keywords such as ``right angle''. This indicates a failure of logical gating: theorem invocation is driven by lexical heuristics rather than condition verification. In multi-step reasoning, such incorrect activation propagates errors, resulting in premise oversight where the reasoning path is syntactically plausible but logically misaligned with the underlying constraints.

\subsection{Conclusion Misassignment to Entities}

\begin{figure}[htbp] 
  \centering      
  \includegraphics[width=1.0\linewidth]{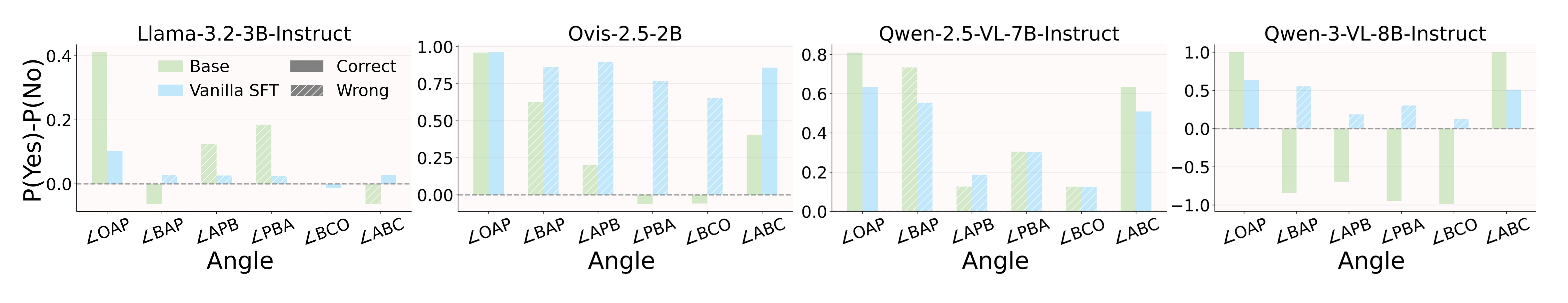} 
\caption{
Object referential error on a geometric right-angle identification task. Here, $O$ is the circle center, $AC$ is a diameter, $B$ lies on the circumference, and $AP$ is the tangent at $A$. The model predicts whether a queried angle is $90^\circ$. The x-axis shows different candidate angles, and the y-axis shows $P(\text{Yes}) - P(\text{No})$. Although only $\angle OAP$ and $\angle ABC$ are valid right angles, baseline models already over-rely on keywords such as ``tangent'' and ``circle'', while vanilla SFT further exacerbates conclusion misassignment.
}
  \label{fig:AngleReason}
\end{figure}

We define \textbf{Conclusion Misassignment to Entities} as the failure to correctly bind theorem-level conclusions to their corresponding problem-specific entities. Even when the correct rule is selected, models often fail to align the theorem’s abstract structure with the underlying geometric configuration.
As shown in Figure~\ref{fig:AngleReason}, the model may recognize patterns such as ``tangent-perpendicular'', but cannot reliably instantiate abstract roles (e.g., $[Radius]$, $[Tangent]$) with concrete entities (e.g., $OA$, $AP$). Consequently, models frequently assign the $90^\circ$ property to incorrect entities such as $\angle BAP$ or $\angle BCO$.
This suggests that reasoning is not structurally grounded, but instead behaves like bag-of-entities classification driven by surface associations. Vanilla SFT further exacerbates this issue by encouraging shortcut learning from end-to-end supervision, reinforcing weak keyword-property correlations (e.g., ``tangent'' $\rightarrow$ $90^\circ$) instead of robust variable-role binding. As a result, the model fails to preserve invariant entity mappings under changes in labels or geometric configurations.

%% file: sec/4_TheoremLearning.tex
\section{Theorem-SFT: Reorienting Supervision toward Structural Rules}
\label{sec:theorem-sft}


To bridge the generalization gap identified in Sec.~\ref{sec:whyFail}, we propose \textbf{Theorem-SFT}. The core philosophy of Theorem-SFT is a shift in the \textbf{inductive target}: from the imitation of problem-specific trajectories to the acquisition of invariant structural rules. As discussed in Sec.~\ref{sec:intro}, human learners do not master reasoning by memorizing thousands of specific instances; instead, they internalize abstract theorems that can be flexibly instantiated across diverse contexts. Theorem-SFT operationalizes this by decoupling logical principles from problem-specific noise.

\begin{figure}[htbp] 
  \centering      
  \includegraphics[width=0.9\linewidth]{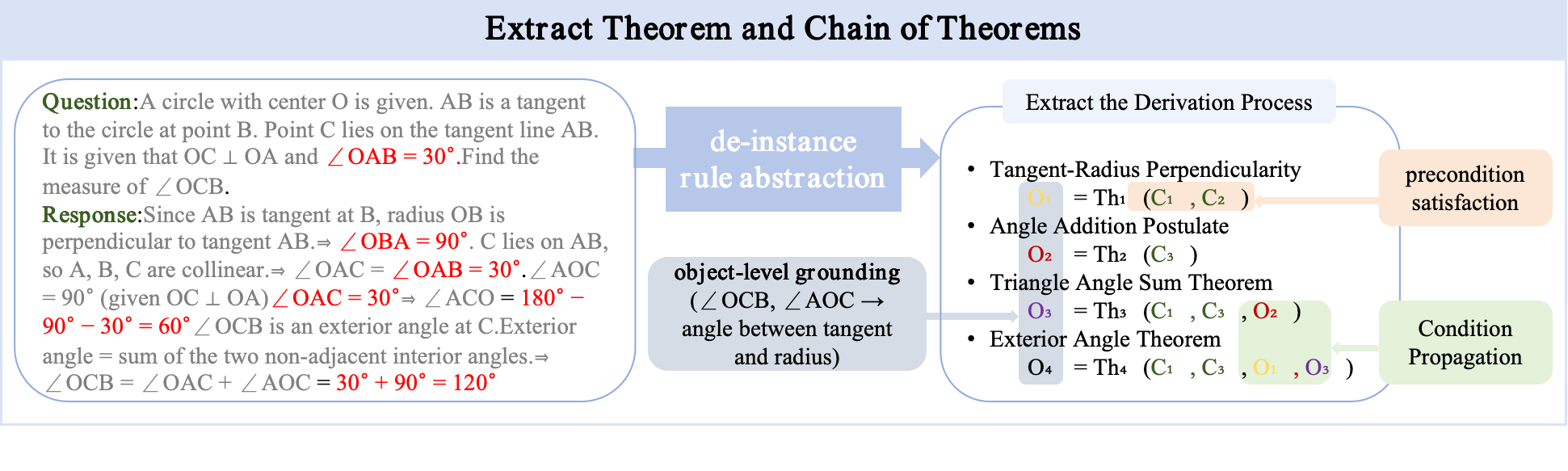} 
  \caption{
Illustration of theorem extraction and theorem-chain construction from problem-solution pairs.
}
  \label{fig:TheoremLearning}
\end{figure}

As illustrated in Figure~\ref{fig:TheoremLearning}, the Theorem-SFT pipeline operates in two progressive stages. First, \textbf{Theorem Learning} (Sec.~\ref{sec:theo_learn}) decomposes individual theorems into structured functional primitives---explicitly modeling their logical preconditions and entity-role mappings to ensure single-step precision. Second, \textbf{Chain-of-Theorems} (Sec.~\ref{sec:chain_learn}) synthesizes these atomic rules into deductive sequences, focusing on how conditions propagate across steps to ensure multi-step consistency. By decoupling logical principles from problem-specific noise, Theorem-SFT enables the model to transition from surface-level pattern matching to systematic rule execution, mirroring the deductive process of human mathematical reasoning.



\subsection{Theorem Learning: Learning Atomic Rules via De-instantiation}

\label{sec:theo_learn}The first stage of Theorem-SFT focuses on De-instantiation, the process of decoupling abstract logical principles from the noise of specific problem instances. Reflecting human pedagogy, where a student must master the applicability and structural roles of a rule before invoking it, we formalize each theorem $\mathcal{T}$ as a functional operator:\begin{equation}\mathcal{T} = <{ \text{Cond}, \text{Map}, \text{Exec} }>\end{equation}By reformulating theorem knowledge into this structured format (see Fig.~\ref{fig:datacompare}), we explicitly supervise the model across three functional dimensions:

\begin{itemize}
\item \textbf{Conditions as Logical Gating ($\text{Cond}$):} A hallmark of human expertise is the ``sanity check'' performed before theorem invocation. While vanilla SFT models often suffer from "hallucinated triggers" based on superficial keywords, Theorem-SFT mandates an explicit verification of the Conditions field. By incorporating Counterexamples as negative supervision, the model learns the logical boundary of the rule. This trains the model to recognize when a theorem's premises are \textit{not} met, effectively curbing the \textit{Premise Oversight} identified in Sec.~\ref{sec:whyFail}.

\item \textbf{Entity Mapping for Structural Binding ($\text{Map}$):} Human reasoning excels at mapping abstract roles to concrete entities regardless of their specific labels. We introduce a dedicated Entity Mapping layer to simulate this referential alignment. This component requires the model to bind the theorem's abstract variables (e.g., $[base]$, $[height]$) to problem-specific objects (e.g., segment $BC$, altitude $AD$). By explicitly supervising this binding process, we ensure the model treats reasoning as a structured assignment task, thereby resolving \textit{Conclusion Misalignment} failure.

\item \textbf{Intuition for Deductive Execution ($\text{Exec}$):} The Intuition field provides an instance-independent description of the rule's execution logic. Once the logical gate ($\text{Cond}$) is verified and the entities are bound ($\text{Map}$), the model executes the invariant logical operator ($\text{Exec}$) to derive the conclusion. This mirrors the human ability to generalize a single concept across infinite problem variations by focusing on the underlying rule rather than the textual surface.
\end{itemize}

In short, this stage transforms the model from a probabilistic pattern matcher that mimics solution surfaces into a structural solver that executes verified rules. By enforcing the $\text{Cond} \to \text{Map} \to \text{Exec}$ sequence, Theorem-SFT ensures that every step of reasoning is grounded in logical necessity and precise referential mapping.

\subsection{Chain-of-Theorems: Compositional Reasoning via Deductive Theorem Chains}\label{sec:chain_learn}

Proficiency in isolated rules is a prerequisite for, but not the entirety of, complex reasoning. Another huge challenge lies in compositional deduction---weaving multiple atomic theorems into a coherent, hierarchical proof. Mirroring the human deductive process, where a complex problem is decomposed into a sequence of interdependent steps, we introduce \textbf{Chain-of-Theorems}. This stage trains the model to treat reasoning not as a textual sequence, but as a structured assembly of functional blocks (see the bottom row of Figure~\ref{fig:datacompare}). As illustrated in our data samples, we explicitly model the composition of rules through three progressive layers:

\begin{itemize}
\item \textbf{Source Theorem Identification:} Before executing a derivation, the model is supervised to identify the set of Source Theorems (e.g., \textit{Alternate Interior Angles Theorem}, \textit{Linear Transformation Rules}) required for the task. This step simulates the human cognitive process of tool selection, ensuring that the model recognizes the atomic primitives $\mathcal{T}_i$ needed to build a complex chain.

\item \textbf{Step-by-Step Theorem Composition:} This field explicitly models the handover of logical states across a trajectory. As shown in the geometry and algebra samples, the Theorem Composition partitions the proof into discrete stages (Step 1 to Step N). This ensures \textit{Condition Propagation}, where the conclusion $(\text{Exec})$ of a preceding step is strictly verified to satisfy the premises $(\text{Cond})$ of the subsequent step. This transforms reasoning from flat text into a directed logical flow, effectively resolving the consistency issues common in multi-step SFT.

\item \textbf{High-Order Abstraction (Chain Definition):} To reinforce the structural integrity of the reasoning, we provide a Definition of the chain and its Intuition. This component requires the model to summarize the composite logic as a new, higher-order functional form. By training on both natural language and formal versions of these chains, the model learns to internalize the invariant reasoning architecture that persists across varied problem instances.
\end{itemize}

 In essence, this stage elevates the model from ``sentence-by-sentence generation'' to ``hierarchical functional composition''. By learning to synthesize atomic rules into verified Theorem Chains, the model acquires the ability to construct deep reasoning trajectories that are grounded in structural dependency and cross-step consistency rather than surface-level textual fluency.

\subsection{Data Corpus and Statistics}
\begin{figure}[htbp] 
  \centering      
  \includegraphics[width=0.9\linewidth]{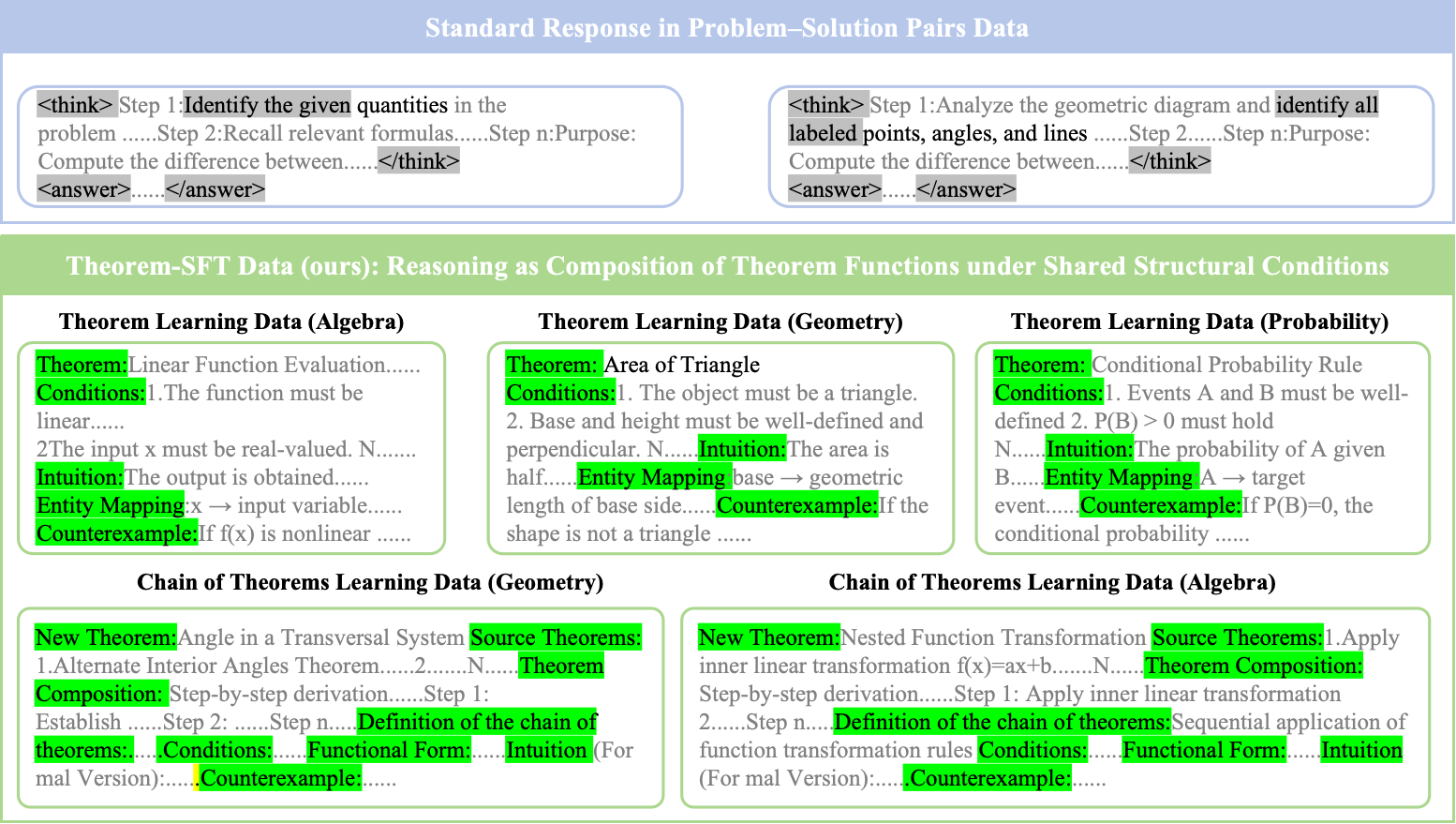} 
  \caption{
This figure compares standard responses in problem-solution datasets and Theorem-SFT data. We highlight key words to emphasize their structural differences. Theorem-SFT contains two structures for theorem learning and theorem-chain learning. The consistent format across algebra, geometry, and probability suggests that our formulation generalizes beyond specific mathematical domains.
}
  \label{fig:datacompare}
\end{figure}

To ensure a rigorous evaluation of Theorem-SFT, we construct a large-scale structural supervision dataset based on the MATH and GeoQA training corpora. Our construction process yields a diverse set of reasoning primitives and deductive chains across multiple mathematical disciplines.\begin{itemize}\item \textbf{Atomic Theorem Corpus:} We extract and formalize approximately \textbf{3.8K} geometric theorems from GeoQA and \textbf{6.4K} algebraic and probabilistic principles from MATH. As shown in Figure~\ref{fig:datacompare}, each entry is meticulously structured into the $\{\text{Cond}, \text{Map}, \text{Exec}\}$ format.\item \textbf{Theorem Chain Diversity:} For compositional reasoning, we synthesize over \textbf{8K} deductive chains. These chains vary in depth, ranging from 2-step basic compositions to complex 5-step hierarchical proofs, ensuring the model encounters various logical propagation patterns.
\end{itemize}

%% file: sec/5_Results.tex
\section{Experiments}\label{sec:exe}
\subsection{Settings}
\textbf{Datasets and Evaluation.} We evaluate our method on two distinct reasoning domains: (i) General Mathematical Reasoning, using the MATH-500 test set and competition-level benchmarks including AGIEval-Math~(\cite{zhong2024agieval}), Omni-Math~(\cite{gao2024omni}), and LiveBench-Math~(\cite{white2024livebench}); and (ii) Visual-Spatial Reasoning, using the GeoQA~(\cite{chen2021geoqa}) multimodal benchmark. To test the boundary of our method, we also include GSM1K~(\cite{zhang2024careful}), which focuses on basic arithmetic rather than complex theorem application.

\textbf{Training Parity and Knowledge Source.} To ensure a strictly controlled comparison, all theorem annotations used for Theorem-SFT are extracted exclusively from the original training sets of MATH and GeoQA. The GeoQA theorem corpus comprises approximately 3.8K structural rules, while the MATH corpus includes 6.4K. Crucially, Theorem-SFT and Vanilla SFT share the identical underlying knowledge source, with Theorem-SFT merely re-organizing the training signals into structural functional blocks $\{\text{Cond, Map, Exec}\}$.

\textbf{Baseline Paradigm.} We compare Theorem-SFT against two baselines: (1) Baseline Models, which represent the pre-trained/instruct-tuned capability without additional fine-tuning on the target theorem set; (2) Vanilla SFT, which follows the standard paradigm of training on problem-solution pairs. By evaluating on held-out test sets that models have never encountered during training, any performance gain can be definitively attributed to the superior generalization of de-instantiated rules rather than the memorization of reasoning trajectories.

\textbf{Implementation Details.} We conduct experiments across multiple model families and scales, including LLaMA-3.2 (3B/8B)~(\cite{grattafiori2024llama}), Mistral-7B~(\cite{llmjiang2023mistral7b}), and the vision-language Qwen2.5-VL (3B/7B) and Gemma-3-4B. All models are fine-tuned using LoRA~\cite{hu2022lora} with two variants: \textbf{T-All} (tuning all layers) and T-MLP (tuning only MLP layers). Training is performed with a constant learning rate of \textbf{1e-5} and a total batch size of \textbf{32}.

\subsection{Performance}
\label{sec:results}

\subsubsection{Language Model Results}
Table~\ref{tab:math_llm_results} shows that Theorem-SFT consistently improves performance on MATH-500 across model scales, with gains of +8.8\% (LLaMA-3.2-3B-Instruct), +4.8\% (LLaMA-3.2-8B-Instruct), and +14.4\% (Mistral-7B-Instruct). Improvements extend to other benchmarks, and are more pronounced on complex reasoning tasks, e.g., +8.6\% and +7.1\% on AGIEval-Math for the two LLaMA models. Gains on simpler datasets like GSM1K are modest ($\sim$1\%), indicating that Theorem-SFT is especially effective for structured, multi-step reasoning.

Notably, although trained solely on theorems from the MATH training set, the model generalizes robustly to unseen competition-level datasets such as AGIEval-Math, LiveBench-Math, and Omni-Math. This cross-benchmark transfer suggests that these datasets share an underlying space of mathematical concepts and theorem usage. By learning de-instantiated theorem representations, Theorem-SFT captures this shared structure. Gains are consistent across 3B and 8B scales, indicating that structural supervision provides a universal inductive bias independent of model capacity.

Overall, these findings demonstrate that theorem-level supervision provides a strong inductive bias for generalizable reasoning. In particular, it enables models to learn how to apply abstract mathematical principles rather than memorize problem-solution patterns, leading to significantly better transfer across datasets. This highlights that theorem-level supervision is more generalizable than conventional problem-answer supervision.

\begin{table*}[ht]
\centering
\scriptsize
\setlength{\tabcolsep}{5.8pt}

\begin{tabular}{l l cccccc}
\hline
Model & Method 
& MATH-500 & GSM1K & AGIEval-SAT & AGIEval-Math & LiveBench & Omni-Math \\
\hline

\multirow{4}{*}{LLaMA-3.2-3B-Instruct}
& Base        & 30.6 & 56.3 & 58.6 & 27.6 & 35.6 & 11.6 \\
& Vanilla SFT & 23.4 & 39.0 & 57.3 & 25.1 & \textbf{45.4} & 5.2 \\
& \hl{\textbf{T-All}}    
& \hl{\textbf{39.4}} & \hl{57.2} & \hl{65.0} & \hl{35.8} & \hl{42.4} & \hl{\textbf{13.6}} \\
& \hl{\textbf{T-MLP}}    
& \hl{38.0} & \hl{\textbf{57.5}} & \hl{\textbf{67.3}} & \hl{\textbf{36.2}} & \hl{44.6} & \hl{11.9} \\
\hline

\multirow{4}{*}{LLaMA-3.1-8B-Instruct}
& Base        & 40.4 & 73.9 & 72.3 & 38.5 & 47.9 & 17.3 \\
& Vanilla SFT & 35.6 & 64.2 & 65.5 & 33.1 & 47.9 & 17.1 \\
& \hl{\textbf{T-All}}    
& \hl{\textbf{45.2}} & \hl{\textbf{76.1}} & \hl{73.6} & \hl{\textbf{45.6}} & \hl{52.3} & \hl{18.1} \\
& \hl{\textbf{T-MLP}}    
& \hl{43.0} & \hl{75.4} & \hl{\textbf{78.2}} & \hl{45.1} & \hl{\textbf{53.6}} & \hl{\textbf{18.4}} \\
\hline

\multirow{4}{*}{Mistral-7B-Instruct}
& Base        & 40.6 & 39.9 & 43.2 & 13.0 & \textbf{39.7} & 7.3 \\
& Vanilla SFT & 44.4 & 26.7 & 38.2 & 15.1 & 31.0 & 6.3 \\
& \hl{\textbf{T-All}}    
& \hl{\textbf{55.0}} & \hl{\textbf{41.4}} & \hl{50.3} & \hl{\textbf{19.3}} & \hl{37.6} & \hl{\textbf{9.2}} \\
& \hl{\textbf{T-MLP}}    
& \hl{54.0} & \hl{40.8} & \hl{\textbf{51.8}} & \hl{18.5} & \hl{38.0} & \hl{9.1} \\
\hline

\end{tabular}

\caption{
Performance (\%) on mathematical benchmarks for three LLMs. 
T-All and T-MLP refer to our proposed theorem-supervised (Theorem-SFT) methods. 
T-All applies theorem-level supervision to all layers, 
while T-MLP applies it only to MLP layers.
}
\label{tab:math_llm_results}
\end{table*}

\subsubsection{Cross-modal Generalization}
On the GeoQA~(\cite{chen2021geoqa}) vision-language benchmark, Theorem-SFT achieves consistent improvements across different model scales. Specifically, we observe gains of +5.94\% on Qwen2.5-VL-3B-Instruct~(\cite{qwen2025vl}), +20.27\% on Qwen2.5-VL-7B-Instruct, and +13.48\% on Gemma-3-4B-IT~(\cite{llmsellergren2025medgemma}).

Importantly, all models are trained exclusively with text-based theorem supervision but evaluated on GeoQA, a multimodal benchmark requiring visual-spatial reasoning. Despite this cross-modal gap, all models show consistent improvements, with Qwen2.5-VL-7B-Instruct achieving the largest gain of +20.27\%.

These results suggest that rule-based knowledge exhibits strong cross-modal transferability: although learned in textual form, it generalizes effectively to visual reasoning tasks. This implies a potential training paradigm in which learning abstract rules in text provides a foundation for multimodal reasoning.

\begin{table*}[ht]
\centering
\scriptsize
\begin{tabular}{lcccc|cccc|cccc}
\hline
& \multicolumn{4}{c}{Qwen2.5-VL-3B}
& \multicolumn{4}{c}{Qwen2.5-VL-7B}
& \multicolumn{4}{c}{Gemma-3-4B-IT} \\
\cline{2-13}
Dataset
& Base & Vanilla & T-All & T-MLP
& Base & Vanilla & T-All & T-MLP
& Base & Vanilla & T-All & T-MLP \\
\hline

GeoQA
& 35.95 & 34.59 & 35.54 & \textbf{41.89}
& 49.46 & 38.78 & 66.89 & \textbf{69.73}
& 23.24 & 31.89 & 35.72 & \textbf{36.72} \\
\hline
\end{tabular}
\caption{Performance (\%) on GeoQA benchmark. T-All and T-MLP refer to our proposed theorem-supervised (Theorem-SFT) methods. T-All denotes Theorem-SFT applied to all layers, while T-MLP applies it only to MLP layers.}
\label{tab:geoqa_results}
\end{table*}

\subsection{Investigations}

Prior work~(\cite{geva2021transformer,meng2022locating,dai2022knowledge}) on large language models suggests that MLP (feed-forward network, FFN) layers are more closely associated with storing factual and parametric knowledge, while attention layers primarily model contextual interactions.

\begin{figure}[htbp] 
  \centering      
  \includegraphics[width=1.0\linewidth]{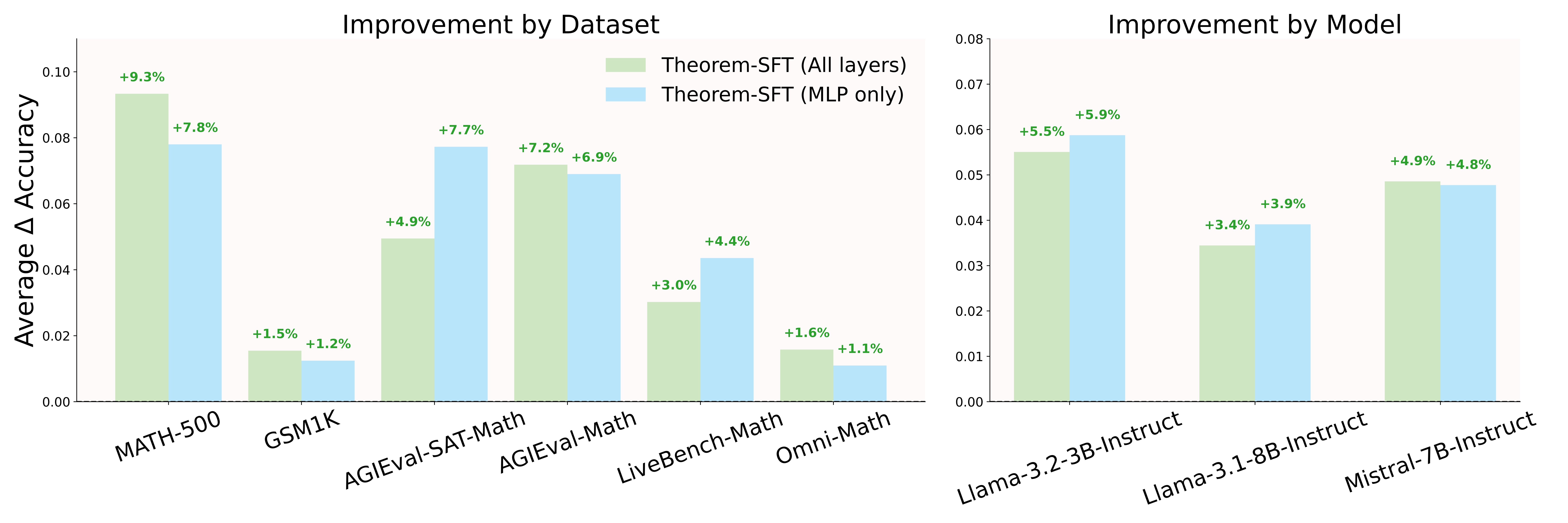} 
\caption{Where are theorems stored? Evidence from parameter-efficient tuning. The figure compares full-layer and MLP-only LoRA tuning for Theorem-SFT. (Left) Average accuracy improvements across six mathematical reasoning benchmarks. (Right) Average improvements across three base models.}
  \label{fig:MLP_theorem}
\end{figure}

To investigate where theorem-level knowledge is encoded, we compare two variants of Theorem-SFT using LoRA~(\cite{hu2022lora}): (i) updating all layers, and (ii) updating only MLP layers. As shown in Figure~\ref{fig:MLP_theorem}, MLP-only adaptation achieves performance comparable to full-model fine-tuning across most benchmarks. Notably, on multimodal GeoQA tasks, MLP-only tuning even slightly outperforms full-layer adaptation.

These results suggest that MLP layers may play a more important role in encoding theorem-related knowledge, rather than attention-based contextual interactions. One possible explanation is that such knowledge is stored in a parametric form, primarily encoded within feed-forward transformations.
This finding further implies that improving reasoning generalization depends less on which layers are tuned, and more on where structured knowledge is encoded within model parameters.

Additional analyses of theorem-application errors, weight changes and prompt-side hidden-state drift, and accuracy versus response length are provided in Appendices~\ref{sec:supp_errors}, \ref{sec:internal_details}, and~\ref{sec:efficiency}, respectively.


%% file: sec/2_relatedwork.tex
\section{Related Work}

\paragraph{Supervised Fine-Tuning for Reasoning}
Supervised fine-tuning (SFT) is widely used to boost reasoning in LLMs by training on step-by-step demonstrations (\cite{ouyang2022training,wei2021finetuned,wang2023self,lu2023llama,bolton2024biomedlm,singhal2025toward,guha2023legalbench,llmachiam2023gpt,llmbai2025qwen3,llmbrown2020language,llmjiang2023mistral7b,llmlu2025ovis2,llmsellergren2025medgemma,llmtouvron2023llama}). Yet prior work shows it often picks up surface patterns instead of transferable reasoning principles, leading to poor generalization (\cite{chu2025sft,chen2025sft}). While this performance gap is well documented, what kind of knowledge SFT actually acquires remains underexplored.

\paragraph{Memorization and Generalization in LLMs and VLMs}
Research on memorization and generalization reveals that supervised training frequently latches onto spurious cues like lexical patterns and dataset quirks, hurting out-of-distribution performance (\cite{han2022images,yang2023resmem,carlini2022quantifying,bandel2022lexical,serrano2023stubborn,du2022less}). Most analyses focus on whether or how much memorization happens, but rarely dig into what gets memorized---especially whether structured knowledge like rules, constraints, and applicability conditions is actually learned.

\paragraph{Mathematical Reasoning}
Mathematical reasoning has been heavily benchmarked via MATH, GSM8K, GeoQA, AGIEval, LiveBench-Math, and Omni-Math (\cite{lightman2023let,zhang2024careful,chen2021geoqa,zhong2024agieval,white2024livebench,gao2024omni}). Chain-of-thought prompting boosts multi-step reasoning by eliciting intermediate steps (\cite{wei2022chain}). However, models fine-tuned on CoT data via SFT often do well in-distribution but fall apart on unseen or compositional problems (\cite{kang2025learnin}). This hints that current methods miss the structured, reusable nature of mathematical reasoning.

%% file: sec/6_conclusion.tex
\section{Conclusion and Limitations}
\textbf{Conclusion.} We revisit SFT's generalization failures in mathematical reasoning and argue the root cause is not memorization, but the failure to internalize structural invariants of logical rules, the gating of preconditions and binding of entity roles. We propose Theorem-SFT, which shifts supervision from trajectory imitation to structural learning via De-instantiation. By decomposing theorems into functional blocks (\(\text{Cond}, \text{Map}, \text{Exec}\)) and supervising their composition, we mitigate \textit{Premise Oversight} and \textit{Conclusion Misassignment}. Theorem-SFT yields consistent gains across benchmarks and scales without direct problem-solution supervision, suggesting reasoning generalization benefits from grounding parameters in functional logical structure rather than instance accumulation.

\textbf{Limitations.}\label{sec:limitation} We mainly evaluate mathematical domains, leaving its generalization to broader reasoning tasks, interaction with RL, and scaling behavior on larger theorem corpora as open questions. Error counts measure occurrences rather than the fraction of incorrect answers and can depend on response length. Internal measurements cover one checkpoint and a fixed prompt set; they do not directly establish knowledge retention or a causal mechanism for generalization.

%% file: sec/7_appendix.tex
\section{Technical appendices and supplementary material}
\subsection{Model Responses}
\subsubsection{Case 1: Angle Reasoning with Incorrect Theorem Application}
\paragraph{Question}
$\overline{BC}$ is parallel to the segment through $A$, and $AB = BC$. What is the number of degrees represented by $x$?

\begin{center}

draw((0,0)--(10,0));
draw((0,3)--(10,3));
draw((2,3)--(8,0));
draw((2,3)--(4,0));
label("$A$",(2,3),N);
label("$B$",(4,0),S);
label("$C$",(8,0),S);
label("$124^{\circ}$",(2,3),SW);
label("$x^{\circ}$",(4.5,3),S);

\end{center}

\textbf{Ground Truth:} $x = 28^\circ$

\vspace{0.5em}
\hrule
\vspace{0.5em}

\paragraph{Baseline}
\textbf{Answer:} $56^\circ$ \\
\textbf{Correct:} $\times$

<think>
1.  $\overline{BC}$ is parallel to the segment through $A$, so $\angle BAC \cong \angle ABC$ (Corresponding Angles Postulate).
2.  Since $AB = BC$, $\triangle ABC$ is an isosceles triangle.
3.  In an isosceles triangle, the base angles are congruent, so $\angle BAC \cong \angle ACB$ (Isosceles Triangle Theorem).
4.  The sum of the interior angles of a triangle is $180^\circ$, so $m\angle BAC + m\angle ABC + m\angle ACB = 180^\circ$ (Angle Sum Theorem).
5.  Since $\angle BAC \cong \angle ABC$ and $\angle BAC \cong \angle ACB$, we have $2m\angle BAC + m\angle BAC = 180^\circ$.
6.  Simplifying, we get $3m\angle BAC = 180^\circ$.
7.  Solving for $m\angle BAC$, we have $m\angle BAC = \frac{180^\circ}{3} = 60^\circ$.
8.  Since $\angle BAC \cong \angle ABC$, we have $m\angle ABC = 60^\circ$.
9.  We are given that $m\angle BAC = 124^\circ$, so we can find $x$ by subtracting $m\angle BAC$ from $180^\circ$: $x = 180^\circ - 124^\circ = 56^\circ$.
</think>
<answer>
56
</answer>

\vspace{0.5em}
\hrule
\vspace{0.5em}

\paragraph{Vanilla SFT}
\textbf{Answer:} $56^\circ$ \\
\textbf{Correct:} $\times$

<think>
Since $\overline{BC}$ is parallel to the segment through $A$, then $\angle ABC$ and $\angle BAC$ are corresponding angles.  By the Corresponding Angles Theorem, they are congruent.  Thus, $\angle ABC = \angle BAC = 124^{\circ}/2 = 62^{\circ}$.  Since $AB = BC$, then $\triangle ABC$ is isosceles, so $\angle BCA = 180^{\circ} - 62^{\circ} - 62^{\circ} = \boxed{56^{\circ}}$.
</think><answer>56</answer>

\vspace{0.5em}
\hrule
\vspace{0.5em}

\paragraph{Theorem-SFT}
\textbf{Answer:} $28^\circ$ \\
\textbf{Correct:} $\checkmark$

<think>
1. Since $\overline{BC}$ is parallel to the segment through $A$, corresponding angles are equal (Corresponding Angles Theorem).
2. Let the angle at $A$ be $\angle BAD$ and the angle at $B$ be $\angle BAC$.
3. Since $AB = BC$, $\triangle ABC$ is isosceles (SAS).
4. $\angle BAC = \angle ABC$ (Isosceles Triangle Theorem).
5. The sum of interior angles in $\triangle ABC$ is $180^{\circ}$ (Angle Sum Property of a Triangle).
6. $\angle BAD + \angle BAC + \angle ABC = 180^{\circ}$.
7. Substituting $\angle BAD = 124^{\circ}$ and $\angle BAC = \angle ABC = x^{\circ}$ into the equation.
8. $124^{\circ} + x^{\circ} + x^{\circ} = 180^{\circ}$.
9. Simplifying the equation gives $2x^{\circ} + 124^{\circ} = 180^{\circ}$.
10. Subtracting $124^{\circ}$ from both sides gives $2x^{\circ} = 56^{\circ}$.
11. Dividing by $2$ gives $x^{\circ} = 28^{\circ}$.
</think>
<answer>
28
</answer>

\subsubsection{Case 2: Cubic Equation Reasoning}

\paragraph{Question}
Find the roots of $(x - 3)^3 + (x -7)^3 = (2x - 10)^3.$

\textbf{Ground Truth:} $x = 3, 5, 7$

\vspace{0.5em}
\hrule
\vspace{0.5em}

\paragraph{Baseline}
\textbf{Answer:} $x = 3, -15, 1$ \\
\textbf{Correct:} $\times$

<think>
1. Expand the equation using the formula $(a - b)^3 = a^3 - 3a^2b + 3ab^2 - b^3$.
2. Apply the formula to all terms: $x^3 - 9x^2 + 27x - 27 + x^3 - 21x^2 + 63x - 343 = 8x^3 - 60x^2 + 160x - 1000$.
3. Combine like terms: $2x^3 - 30x^2 + 90x - 1270 = 8x^3 - 60x^2 + 160x - 1000$.
4. Move all terms to one side of the equation: $6x^3 + 30x^2 - 70x + 270 = 0$.
5. Factor out the greatest common factor of $6$: $6(x^3 + 5x^2 - \frac{35}{3}x + 45) = 0$.
6. Notice that the equation is a cubic equation, so we need to find a rational root using the Rational Root Theorem (Theorem 1).
7. The Rational Root Theorem states that any rational root must be of the form $\frac{p}{q}$, where $p$ is a factor of the constant term and $q$ is a factor of the leading coefficient.
8. In this case, the constant term is 45 and the leading coefficient is 1, so the possible rational roots are the factors of 45.
9. Test the possible rational roots using synthetic division or direct substitution to find that $x = 3$ is a root.
10. Use polynomial division or synthetic division to divide the cubic polynomial by $(x - 3)$ to find the remaining quadratic factor.
11. The remaining quadratic factor is $x^2 + 8x - 15$.
12. Factor the quadratic expression using the quadratic formula (Formula 2) or by finding two numbers that multiply to -15 and add up to 8.
13. The quadratic expression factors as $(x + 15)(x - 1)$.
14. Therefore, the roots of the original equation are $x = 3, x = -15, x = 1$.

Formulas and Theorems used:
(1) The Rational Root Theorem: If a rational number $p/q$ is a root of the polynomial $a_nx^n + a_{n-1}x^{n-1} + \ldots + a_1x + a_0$, then $p$ is a factor of $a_0$ and $q$ is a factor of $a_n$.
(2) The Quadratic Formula: If $ax^2 + bx + c = 0$, then $x = \frac{-b \pm \sqrt{b^2 - 4ac}}{2a}$.

Note: The final answer is not requested in the output format.
</think>

\vspace{0.5em}
\hrule
\vspace{0.5em}

\paragraph{Vanilla SFT}
\textbf{Answer:} $x = 6$ \\
\textbf{Correct:} $\times$

<think>
Let $y = x - 5.$ Then $x = y + 5.$
Substituting, we get $(y + 8)^3 + (y - 2)^3 = (3y)^3.$
By the sum of cubes formula, $y + 8 + y - 2 = 27y.$
Solving, we get $y = 1.$
Then $x = y + 5 = 6.$
</think>
<answer>6</answer>

\vspace{0.5em}
\hrule
\vspace{0.5em}

\paragraph{Theorem-SFT}
\textbf{Answer:} $x = 3, 5, 7$ \\
\textbf{Correct:} $\checkmark$

<think>
1. Expand the equation using the identity $(a - b)^3 = a^3 - 3a^2b + 3ab^2 - b^3$.
2. Expand the left-hand side: $(x - 3)^3 = x^3 - 9x^2 + 27x - 27$ and $(x - 7)^3 = x^3 - 21x^2 + 147x - 343$.
3. Expand the right-hand side: $(2x - 10)^3 = 8x^3 - 120x^2 + 400x - 1000$.
4. Combine like terms: $x^3 - 9x^2 + 27x - 27 + x^3 - 21x^2 + 147x - 343 = 8x^3 - 120x^2 + 400x - 1000$.
5. Simplify the equation: $2x^3 - 30x^2 + 174x - 370 = 8x^3 - 120x^2 + 400x - 1000$.
6. Move all terms to one side: $0 = 6x^3 - 90x^2 + 226x - 630$.
7. Factor out common terms: $0 = 6(x^3 - 15x^2 + 37.67x - 105)$.
8. Apply the Rational Root Theorem to find possible roots.
9. Check if $x = 5$ is a root (by inspection or synthetic division).
10. Factor the cubic polynomial: $0 = 6(x - 5)(x^2 - 10x + 21)$.
11. Solve the quadratic equation $x^2 - 10x + 21 = 0$ using the quadratic formula: $x = \frac{-b \pm \sqrt{b^2 - 4ac}}{2a}$.
12. Apply the quadratic formula: $x = \frac{10 \pm \sqrt{(-10)^2 - 4(1)(21)}}{2(1)}$.
13. Simplify the expression: $x = \frac{10 \pm \sqrt{100 - 84}}{2}$.
14. Simplify the expression: $x = \frac{10 \pm \sqrt{16}}{2}$.
15. Simplify the expression: $x = \frac{10 \pm 4}{2}$.
16. Solve for both values: $x = \frac{10 + 4}{2}$ or $x = \frac{10 - 4}{2}$.
17. Simplify the expressions: $x = 7$ or $x = 3$.
</think>
<answer>
x = 5, x = 7, x = 3
</answer>

\subsubsection{Case 3: Arithmetic Error under Correct Rule Application}

\paragraph{Question}
Simplify: $\frac{\sqrt{2.5^2-0.7^2}}{2.7-2.5}$.

\textbf{Ground Truth:} $12$

\vspace{0.5em}
\hrule
\vspace{0.5em}

\paragraph{Baseline}
\textbf{Answer:} $5.36$ \\
\textbf{Correct:} $\times$

<think>
1. Apply the difference of squares formula: $a^2 - b^2 = (a - b)(a + b)$.
2. Simplify the expression: $\sqrt{(2.5 - 0.7)(2.5 + 0.7)}$.
3. Calculate the values inside the square root: $\sqrt{(1.8)(3.2)}$.
4. Simplify the numerator: $\sqrt{5.76}$.
5. Calculate the denominator: $2.7 - 2.5 = 0.2$.
6. Apply the quotient rule for square roots: $\frac{\sqrt{a}}{\sqrt{b}} = \sqrt{\frac{a}{b}}$.
7. Simplify the expression: $\sqrt{\frac{5.76}{0.2}}$.
8. Calculate the value inside the square root: $\sqrt{28.8}$.
9. Simplify the expression: $5.36$.
</think>
<answer>5.36</answer>

\vspace{0.5em}
\hrule
\vspace{0.5em}

\paragraph{Vanilla SFT}
\textbf{Answer:} $11.55$ \\
\textbf{Correct:} $\times$

<think>
Use difference of squares:
\[
\frac{\sqrt{2.5^2-0.7^2}}{2.7-2.5}
= \frac{\sqrt{(2.5-0.7)(2.5+0.7)}}{2.7-2.5}
\]
Simplify the denominator: \(2.7-2.5=0.2\)

Simplify the numerator: \(\sqrt{(2.5-0.7)(2.5+0.7)}=\sqrt{1.8 \cdot 3}=\sqrt{5.4}\)

Simplify: \(\frac{\sqrt{5.4}}{0.2} = \frac{\sqrt{5.4}}{\sqrt{0.2^2}}\)

Use the property of square roots: \(\frac{\sqrt{a}}{\sqrt{b}} = \frac{\sqrt{ab}}{b}\)

Simplify: \(\frac{\sqrt{5.4}}{\sqrt{0.2^2}} = \frac{\sqrt{5.4 \cdot 0.2^2}}{0.2^2}\)

Simplify: \(\frac{\sqrt{5.4 \cdot 0.2^2}}{0.2^2} = \frac{\sqrt{0.216}}{0.04}\)

Simplify: \(\frac{\sqrt{0.216}}{0.04} = \frac{0.462}{0.04}\)

Simplify: \(\frac{0.462}{0.04} = 11.55\)
</think>

<answer>11.55</answer>

\vspace{0.5em}
\hrule
\vspace{0.5em}

\paragraph{Theorem-SFT}
\textbf{Answer:} $12$ \\
\textbf{Correct:} $\checkmark$

<think>
1. Evaluate the numerator: 
   \(\sqrt{2.5^2-0.7^2} = \sqrt{(2.5)^2-(0.7)^2}\).
   Apply the difference of squares formula: \(a^2-b^2 = (a+b)(a-b)\).
   \(\sqrt{(2.5)^2-(0.7)^2} = \sqrt{(2.5+0.7)(2.5-0.7)}\).

2. Simplify the numerator: 
   \(\sqrt{(2.5+0.7)(2.5-0.7)} = \sqrt{(3.2)(1.8)}\).

3. Evaluate the denominator: 
   \(2.7-2.5 = 0.2\).

4. Simplify the expression: 
   \(\frac{\sqrt{(3.2)(1.8)}}{0.2} = \frac{\sqrt{5.76}}{0.2}\).

5. Simplify the square root: 
   \(\sqrt{5.76} = \sqrt{5.76} \approx 2.4\).

6. Simplify the final expression: 
   \(\frac{2.4}{0.2} = 12\).
</think>
<answer>12</answer>

\subsubsection{Case 4: Incorrect Polynomial Factorization and Evaluation}

\paragraph{Question}
Let
\[
x^8 + 3x^4 - 4 = p_1(x) p_2(x) \dotsm p_k(x),
\]
where each non-constant polynomial $p_i(x)$ is monic with integer coefficients, and cannot be factored further over the integers. Compute $p_1(1) + p_2(1) + \dots + p_k(1).$

\textbf{Ground Truth:} $10$

\vspace{0.5em}
\hrule
\vspace{0.5em}

\paragraph{Baseline}
\textbf{Answer:} $15$ \\
\textbf{Correct:} $\times$

<think>
1. Apply the Rational Root Theorem to find possible roots of $x^8 + 3x^4 - 4 = 0$.
2. Factor $x^8 + 3x^4 - 4$ as a difference of squares: $(x^4 + 4)(x^4 - 1)$.
3. Factor further using difference of squares: $(x^4 + 4)(x + 1)(x - 1)(x^2 + x \sqrt{3} + 1)(x^2 - x \sqrt{3} + 1)$.
4. Since $x^2 + x \sqrt{3} + 1$ and $x^2 - x \sqrt{3} + 1$ cannot be factored further over the integers, they are the irreducible polynomials.
5. Evaluate each polynomial at $x = 1$: $p_1(1) = 5$, $p_2(1) = 2$, $p_3(1) = 0$, $p_4(1) = 4$, $p_5(1) = 4$.
6. Sum the values of the polynomials: $5 + 2 + 0 + 4 + 4$.
</think>
<answer> 15 </answer>

\vspace{0.5em}
\hrule
\vspace{0.5em}

\paragraph{Vanilla SFT}
\textbf{Answer:} $6$ \\
\textbf{Correct:} $\times$

<think>
By the rational root theorem, any integer roots of $x^8 + 3x^4 - 4$ must divide $4$.  By the factor theorem, $x = 1$ and $x = -1$ are roots, so we can factor $x^8 + 3x^4 - 4$ as $(x^4 - 4)(x^4 + 4)$.  We can factor $x^4 - 4$ as $(x^2 - 2x + 2)(x^2 + 2x + 2)$.  By completing the square, $x^4 + 4 = (x^2 + 2)^2 - 4$, so $x^4 + 4$ is irreducible over the integers.

Therefore, the desired sum is $1 + 1 + 2 + 2 = \boxed{6}$.
</think><answer> 6 </answer>

\vspace{0.5em}
\hrule
\vspace{0.5em}

\paragraph{Theorem-SFT}
\textbf{Answer:} $10$ \\
\textbf{Correct:} $\checkmark$

<think>
1. Factor the polynomial: 
   \[x^8 + 3x^4 - 4 = (x^4 + 4)(x^4 - 1).\]
   By difference of squares: $a^2 - b^2 = (a - b)(a + b)$.

2. Further factorize $x^4 - 1$ and $x^4 + 4$:
   \[x^4 - 1 = (x^2 - 1)(x^2 + 1) = (x - 1)(x + 1)(x^2 + 1).\]
   $x^4 + 4 = (x^2 + 2x + 2)(x^2 - 2x + 2)$ by completing the square.

3. The non-constant polynomials are $(x - 1)$, $(x + 1)$, $(x^2 + 1)$, $(x^2 + 2x + 2)$, and $(x^2 - 2x + 2)$.

4. Evaluate each polynomial at $x = 1$:
   \[(x - 1) \to 1 - 1 = 0,\]
   \[(x + 1) \to 1 + 1 = 2,\]
   \[(x^2 + 1) \to 1^2 + 1 = 2,\]
   \[(x^2 + 2x + 2) \to 1^2 + 2(1) + 2 = 5,\]
   \[(x^2 - 2x + 2) \to 1^2 - 2(1) + 2 = 1.\]

5. Sum the evaluated polynomials: $0 + 2 + 2 + 5 + 1 = 10$.
</think>
<answer>
10
</answer>

\subsection{Supplementary Error Counts}\label{sec:supp_errors}
Table~\ref{tab:errors_full} compares premise oversight and entity misassignment across Base, Vanilla-SFT, and T-All. Each error type is reported as detected occurrences per 100 analyzed answers, counting multiple occurrences within an answer separately. Each paired entry gives premise and entity error counts, respectively. Dataset names follow the main-text results.

\begin{table}[htbp]
\centering
\caption{Error counts per 100 answers. Each pair gives premise / entity errors. T-All denotes all-layer Theorem-SFT LoRA.}
\label{tab:errors_full}
\small
\begin{tabular}{llccc}
\toprule
Model & Dataset & Base & Vanilla-SFT & T-All \\
\midrule
\multirow{4}{*}{LLaMA-3.2-3B}
& AGIEval-SAT & 24.1 / 30.5 & 24.1 / 35.9 & 18.6 / 22.7 \\
& LiveBench & 53.8 / 64.1 & 69.6 / 79.9 & 55.4 / 62.0 \\
& GSM1K & 19.0 / 40.0 & 17.1 / 56.9 & 13.2 / 43.5 \\
& MATH-500 & 45.4 / 46.8 & 48.0 / 58.6 & 43.0 / 42.2 \\
\midrule
\multirow{4}{*}{Mistral-7B}
& AGIEval-SAT & 31.4 / 37.7 & 23.2 / 36.8 & 15.5 / 29.5 \\
& LiveBench & 73.1 / 73.4 & 71.2 / 70.4 & 71.2 / 63.0 \\
& GSM1K & 18.4 / 43.7 & 21.8 / 61.4 & 16.8 / 48.7 \\
& MATH-500 & 66.0 / 76.4 & 57.2 / 59.2 & 56.0 / 61.4 \\
\bottomrule
\end{tabular}
\end{table}

Relative to Vanilla-SFT, T-All reduces premise errors in seven of eight settings and ties in one; entity errors decrease in seven of eight settings. Both error types decrease on all four datasets for LLaMA-3.2-3B. The improvements are not universal: for Mistral-7B, premise errors on LiveBench remain at 71.2 per 100 answers, while entity errors on MATH-500 increase from 59.2 to 61.4. These counts provide behavioral evidence of improved theorem use in most evaluated settings, rather than a uniform reduction in every error type.

\subsection{Weight Changes and Hidden-State Drift}\label{sec:internal_details}
\begin{figure}[htbp]
\centering
\includegraphics[width=\linewidth]{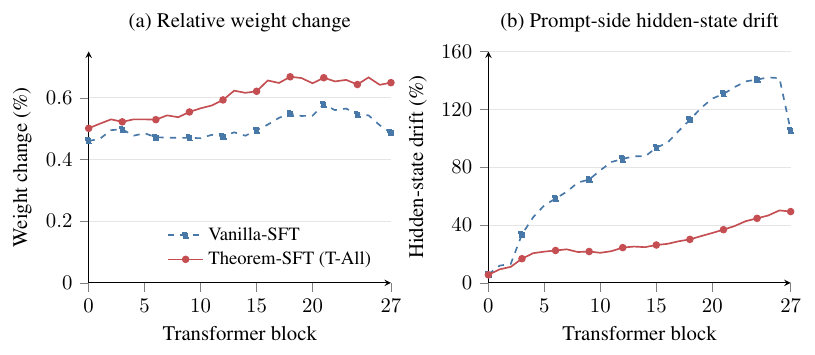}
\caption{Layer-wise changes relative to the base model. (a) Relative weight change normalized by each block's base weight norm. (b) Hidden-state drift at the final prompt token, normalized by the corresponding base hidden-state norm and averaged over the same evaluated prompts for both methods. Lines connect all 28 measurements without smoothing; markers identify every third block. The panels use independent vertical scales. T-All exhibits larger weight changes and smaller prompt-side hidden-state drift in every measured block.}
\label{fig:layerwise_changes}
\end{figure}

We compare parameter changes and prompt-side representation changes for Vanilla-SFT and T-All relative to the same base model. Figure~\ref{fig:layerwise_changes} shows all 28 Transformer blocks of the analyzed checkpoint; Tables~\ref{tab:internal} and~\ref{tab:internal_layers} provide the summary and complete layer-wise measurements.

\paragraph{Measurement.}
For method $m$ and Transformer block $\ell$, let $\mathcal{T}_{\ell}$ denote the parameter tensors in that block. Relative weight change is
\begin{equation}
\mathrm{WC}_{m,\ell}
=\frac{\sqrt{\sum_{i\in\mathcal{T}_{\ell}}\lVert W_{m,i}-W_{0,i}\rVert_F^2}}
{\sqrt{\sum_{i\in\mathcal{T}_{\ell}}\lVert W_{0,i}\rVert_F^2}},
\end{equation}
where $W_{m,i}$ and $W_{0,i}$ are the adapted and base parameter tensors, respectively. Let $h_{m,\ell}(q)$ be the hidden state of the final prompt token for input $q$, and let $Q$ denote the evaluated prompt set. Prompt-side hidden-state drift is
\begin{equation}
\mathrm{HD}_{m,\ell}
=\frac{1}{|Q|}\sum_{q\in Q}
\frac{\lVert h_{m,\ell}(q)-h_{0,\ell}(q)\rVert_2}
{\lVert h_{0,\ell}(q)\rVert_2}.
\end{equation}
Both methods use the same prompts and final prompt-token position. The mean block-wise weight change averages $\mathrm{WC}_{m,\ell}$ across blocks; the aggregated Transformer value applies the same norm ratio jointly to tensors from all blocks. Mean prompt-side drift averages $\mathrm{HD}_{m,\ell}$ across blocks. All reported values below are percentages of the corresponding base-model norms.

\begin{table}[htbp]
\centering
\caption{Summary of internal changes (\% of base-model norms) in the analyzed 28-block checkpoint.}
\label{tab:internal}
\begin{tabular}{lrr}
\toprule
Metric & Vanilla-SFT & T-All \\
\midrule
Mean block-wise relative weight change & 0.504 & 0.598 \\
Aggregated relative weight change & 0.506 & 0.602 \\
Mean prompt-side hidden-state drift & 87.140 & 27.826 \\
\bottomrule
\end{tabular}
\end{table}

\begin{table}[htbp]
\centering
\caption{All 28 Transformer blocks in the internal analysis. W: relative weight change; H: prompt-side hidden-state drift. Values are percentages. V: Vanilla-SFT; T: T-All.}
\label{tab:internal_layers}
\small
\begin{tabular}{rrrrr@{\hspace{16pt}}rrrrr}
\toprule
Block & W(V) & W(T) & H(V) & H(T) & Block & W(V) & W(T) & H(V) & H(T) \\
\midrule
0 & .460 & .501 & 5.877 & 5.608 & 14 & .477 & .616 & 87.463 & 24.711 \\
1 & .467 & .516 & 11.914 & 9.348 & 15 & .495 & .621 & 93.489 & 26.215 \\
2 & .495 & .530 & 13.179 & 11.112 & 16 & .513 & .656 & 96.943 & 26.999 \\
3 & .498 & .522 & 33.605 & 16.743 & 17 & .535 & .648 & 105.117 & 28.783 \\
4 & .477 & .530 & 45.387 & 20.554 & 18 & .549 & .668 & 112.754 & 30.089 \\
5 & .485 & .530 & 53.669 & 21.618 & 19 & .541 & .664 & 120.979 & 32.351 \\
6 & .472 & .529 & 58.291 & 22.402 & 20 & .542 & .647 & 127.437 & 34.525 \\
7 & .471 & .543 & 63.035 & 23.220 & 21 & .579 & .665 & 130.639 & 36.797 \\
8 & .470 & .537 & 69.397 & 21.339 & 22 & .560 & .653 & 135.386 & 39.340 \\
9 & .470 & .554 & 71.433 & 21.694 & 23 & .565 & .658 & 139.360 & 42.718 \\
10 & .469 & .566 & 77.874 & 20.809 & 24 & .546 & .643 & 140.582 & 44.631 \\
11 & .480 & .575 & 83.625 & 21.949 & 25 & .543 & .666 & 142.086 & 46.626 \\
12 & .475 & .593 & 85.675 & 24.400 & 26 & .511 & .642 & 141.561 & 50.125 \\
13 & .488 & .623 & 87.641 & 25.119 & 27 & .486 & .649 & 105.529 & 49.307 \\
\bottomrule
\end{tabular}
\end{table}

\paragraph{Observed contrast.}
T-All exhibits greater relative weight change and smaller prompt-side hidden-state drift than Vanilla-SFT in every measured block. Aggregated weight change is 0.602\% versus 0.506\%, approximately $1.19\times$ as large, while mean prompt-side drift is 27.826\% versus 87.140\%, approximately 31.9\% as large. Thus, larger parameter changes need not imply larger representation changes on the same inputs.

These measurements describe one checkpoint and the evaluated prompts. Prompt-side drift measures pointwise changes in contextualized hidden states, not attention, the complete geometry of the representation space, or knowledge retention directly. The observed contrast complements the behavioral results but does not establish a causal mechanism for generalization.

\subsection{Accuracy and Response Length}\label{sec:efficiency}
Table~\ref{tab:efficiency} reports accuracy and mean response length for LLaMA-3.2-3B using the same answer batches as the main evaluation. Each entry pairs the main-table accuracy with the corresponding mean token count. Theorem-supervised results use T-All.

\begin{table}[htbp]
\centering
\caption{Accuracy (\%) and mean response length (tokens). Each entry is accuracy / tokens.}
\label{tab:efficiency}
\small
\begin{tabular}{lccc}
\toprule
Dataset & Base & Vanilla-SFT & T-All \\
\midrule
AGIEval-Math & 27.6 / 339.1 & 25.1 / 784.6 & 35.8 / 271.2 \\
LiveBench & 35.6 / 670.4 & 45.4 / 1275.1 & 42.4 / 653.4 \\
Omni-Math & 11.6 / 462.5 & 5.2 / 1503.7 & 13.6 / 481.5 \\
\bottomrule
\end{tabular}
\end{table}

T-All produces shorter answers than Vanilla-SFT on all three datasets, with accuracy gains on AGIEval-Math and Omni-Math. On LiveBench, mean length decreases from 1275.1 to 653.4 tokens while accuracy changes from 45.4\% to 42.4\%, illustrating the accuracy--length trade-off. The table characterizes generated-answer length alongside task performance; computational efficiency requires separate latency and cost measurements.

\subsection{Computational Resources}\label{sec:compute}
Experiments use eight NVIDIA H100 GPUs and eight NVIDIA H20 GPUs. We use DeepSpeed ZeRO-2 for memory optimization and FlashAttention for efficient attention computation. All models are trained with bf16 precision. Training is performed with LoRA-based fine-tuning, which significantly reduces computational cost and memory usage. The maximum sequence length is 4096, and gradient accumulation is used to simulate larger batch sizes.

\subsection{Prompts}

In this section, we present the prompts used to construct our theorem-centric supervision data, including theorem extraction, theorem learning, and chain-of-theorems construction. These prompts are designed to enforce structured reasoning and explicit theorem-level supervision.

\subsubsection{Theorem Extraction Prompt}

We design a prompt to extract all relevant mathematical theorems, identities, and problem-solving techniques from a given problem-solution pair.

\textbf{System instruction:}  
You are a mathematics reasoning expert.

\textbf{Task:}  
Extract all mathematical theorems, identities, and problem-solving techniques used in solving the problem.

\textbf{Input:}  
Problem: \{question\}  
Solution: \{response\}

\textbf{Requirements:}  
The model is required to extract only well-established mathematical concepts with standard names. Both explicit and implicit concepts should be included. Each extracted item must be a short canonical name (e.g., “Pythagorean Theorem”, “AM-GM Inequality”). The model must not output descriptive phrases or full sentences, must not invent or paraphrase names, and must exclude concepts without clear standard terminology. Duplicates should be removed and naming should remain consistent.

\textbf{Output format:}  
The output must be valid JSON with a non-empty list under the key “theorems”, and no additional text is allowed.

\textbf{Additional constraints:}  
The model must infer underlying principles even if the solution is short, while avoiding duplication.

\subsubsection{Theorem Learning Prompt}

To construct theorem-level supervision, we design a prompt that enforces a rigorous and formal explanation of each theorem.

\textbf{System instruction:}  
You are a mathematics expert.

\textbf{Task:}  
Provide a rigorous and formal explanation of a given mathematical theorem, identity, or definition.

\textbf{Input:}  
Theorem / Definition: \{theorem\}

\textbf{Requirements:}  
The output must include the following components:

(1) \textbf{Definition:} A precise and formal statement using standard mathematical terminology.

(2) \textbf{Conditions:} All conditions under which the result holds must be explicitly listed, including domains (e.g., integers, real numbers) and constraints (e.g., non-zero, coprime, continuity).

(3) \textbf{Intuition (Formal Version):} The explanation must be mathematically rigorous. All variables and symbols must be explicitly defined (e.g., “Let \(x \in \mathbb{R}\)”, “Assume \(a \neq 0\)”). Informal or visual descriptions are disallowed; instead, all objects (points, segments, angles, arcs) and their relations must be explicitly specified.

(4) \textbf{Examples:} At least two examples must be provided. Each example must define all objects, follow step-by-step reasoning, and clearly indicate where the theorem is applied.

(5) \textbf{Counterexample:} One invalid case must be provided, explicitly stating which condition is violated.

\textbf{Additional constraints:}  
The explanation must be concise, formal, and consistent with standard textbook-style mathematical writing, avoiding unnecessary verbosity.

\subsubsection{Chain-of-Theorems Construction Prompt}

We further design a prompt to construct compositional reasoning structures by deriving new theorems from existing ones.

\textbf{System instruction:}  
You are a mathematics reasoning and theorem discovery researcher.

\textbf{Task:}  
Analyze the reasoning structure of a QA problem, abstract a new derived theorem, explain how it is composed from fundamental theorems, and provide a rigorous formal explanation.

\textbf{Input:}  
Question: \{question\}  
Model Reasoning: \{response\}  
Reference Theorems: \{theorems\} (for inspiration only)

\textbf{Procedure:}

Step 1: Identify all base theorems and definitions that are explicitly used and validly applicable. Missing theorems must not be inferred.

Step 2: Extract structural relations, including equations and constraints. Every transformation must be explicitly justified. Implicit algebraic steps such as division, cancellation, or substitution are not allowed.

Step 3: Derive a new theorem by composing base theorems, only if the derivation is fully valid. The theorem must be general (no specific numerical values) and represent a reusable reasoning pattern. If any step requires unstated assumptions, no theorem should be derived.

Step 4: Explain theorem composition by explicitly showing the logical chain (e.g., Theorem A + Theorem B $\rightarrow$ intermediate result $\rightarrow$ final theorem), with justification at each step and no skipped reasoning.

\textbf{Output structure:}  
The output must include the following components: a concise derived theorem name, source theorems, a detailed theorem composition process, a formal definition, explicit conditions, a functional form (e.g., \(C = f(A, B)\)), a rigorous intuition, two examples with step-by-step reasoning, and one counterexample with violated conditions.

\textbf{Additional constraints:}  
The theorem composition is mandatory and must be explicit. The derived theorem must not be a restatement of an existing theorem. All explanations must be rigorous, with explicitly defined variables and no hidden assumptions or algebraic shortcuts.

\subsection{Theorem Learning Data}

\paragraph{Case 1: Similarity of Polygons}

\textbf{Question.}Given the theorem or definition \textit{Similarity of polygons}, analyze its definition, conditions for application, and key intuition.

\textbf{Definition.} Two polygons are similar if and only if corresponding angles are congruent and corresponding side lengths are proportional.

Let $P_1$ and $P_2$ be two polygons with $n$ sides. Denote their vertices as $A_1, A_2, \dots, A_n$ and $B_1, B_2, \dots, B_n$, respectively. Then $P_1 \sim P_2$ if and only if
\[
\angle A_i = \angle B_i, \quad \forall i = 1, \dots, n,
\]
and
\[
\frac{|A_iA_{i+1}|}{|B_iB_{i+1}|} = k, \quad k > 0, \quad \forall i = 1, \dots, n.
\]

\textbf{Conditions.}
\begin{itemize}
    \item Both polygons have the same number of sides ($n \geq 3$).
    \item Corresponding vertices are consistently ordered.
    \item The ratio of all corresponding sides is a constant $k > 0$.
    \item Polygons are non-degenerate.
\end{itemize}

\textbf{Intuition.} Similarity preserves shape while allowing uniform scaling. Corresponding angles remain invariant, while all side lengths are scaled by a constant factor, ensuring structural equivalence between polygons.

\textbf{Example 1.} Consider triangles with vertices $A_1(0,0)$, $A_2(4,0)$, $A_3(0,3)$ and $B_1(0,0)$, $B_2(8,0)$, $B_3(0,6)$. All corresponding angles are equal and all side ratios equal $1/2$, hence the triangles are similar.

\textbf{Example 2.} Consider squares with vertices $A_1(0,0)$, $A_2(2,0)$, $A_3(2,2)$, $A_4(0,2)$ and $B_1(0,0)$, $B_2(4,0)$, $B_3(4,4)$, $B_4(0,4)$. All angles are $90^\circ$ and all side ratios equal $1/2$, hence they are similar.

\textbf{Counterexample.} Consider triangles with vertices $A_1(0,0)$, $A_2(4,0)$, $A_3(0,3)$ and $B_1(0,0)$, $B_2(8,0)$, $B_3(0,5)$. The corresponding angles are not all equal and side ratios are inconsistent, hence the polygons are not similar.

\paragraph{Case 2: Permutation Cycle Decomposition}

\textbf{Question.} Given the theorem or definition \textit{Permutation Cycle Decomposition}, analyze its definition, conditions for application, and key intuition.

\textbf{Definition.} Let \( S_n \) denote the symmetric group of all permutations on the set \( \{1, 2, \dots, n\} \). A \textbf{permutation cycle decomposition} is a representation of a permutation \( \sigma \in S_n \) as a product of disjoint cycles. Formally, any permutation \( \sigma \in S_n \) can be expressed uniquely (up to the order of the cycles) as:
\[
\sigma = C_1 C_2 \cdots C_k,
\]
where \( C_1, C_2, \dots, C_k \) are disjoint cycles, and each cycle \( C_i \) is a sequence of elements \( (a_1 \, a_2 \, \dots \, a_m) \) such that \( \sigma(a_1) = a_2 \), \( \sigma(a_2) = a_3 \), \dots, \( \sigma(a_m) = a_1 \), and \( \sigma(x) = x \) for all \( x \notin \{a_1, a_2, \dots, a_m\} \).

\textbf{Conditions.}
\begin{itemize}
    \item The domain of \( \sigma \) is the finite set \( \{1, 2, \dots, n\} \), where \( n \in \mathbb{Z}^+ \) (positive integers).
    \item \( \sigma \) must be a bijection, i.e., \( \sigma \in S_n \).
    \item The cycles \( C_1, C_2, \dots, C_k \) must be disjoint, meaning no element of \( \{1, 2, \dots, n\} \) appears in more than one cycle.
    \item The decomposition is unique up to the order of the cycles.
\end{itemize}

\textbf{Intuition.} Let \( \sigma \in S_n \) be a permutation. To understand the cycle decomposition, consider the following formal reasoning:

\begin{enumerate}
    \item \textbf{Definition of a cycle:} A cycle \( (a_1 \, a_2 \, \dots \, a_m) \) describes the action of \( \sigma \) on a subset of elements \( \{a_1, a_2, \dots, a_m\} \subseteq \{1, 2, \dots, n\} \). Specifically:
    \[
    \sigma(a_1) = a_2, \; \sigma(a_2) = a_3, \; \dots, \; \sigma(a_m) = a_1.
    \]
    For any \( x \notin \{a_1, a_2, \dots, a_m\} \), \( \sigma(x) = x \).

    \item \textbf{Disjoint cycles:} Two cycles \( C_i \) and \( C_j \) are disjoint if their supports (the sets of elements they permute) are disjoint. That is, if \( C_i = (a_1 \, a_2 \, \dots \, a_m) \) and \( C_j = (b_1 \, b_2 \, \dots \, b_p) \), then \( \{a_1, a_2, \dots, a_m\} \cap \{b_1, b_2, \dots, b_p\} = \emptyset \).

    \item \textbf{Cycle decomposition process:} Starting from any element \( x \in \{1, 2, \dots, n\} \), repeatedly apply \( \sigma \) to trace the orbit of \( x \) under \( \sigma \). This orbit forms a cycle. Remove these elements from the set \( \{1, 2, \dots, n\} \) and repeat the process with the remaining elements until all elements are exhausted.

    \item \textbf{Uniqueness:} The cycle decomposition is unique up to the order of the cycles because the orbits of \( \sigma \) partition the set \( \{1, 2, \dots, n\} \) into disjoint subsets, and the action of \( \sigma \) within each subset is uniquely determined.
\end{enumerate}

\textbf{Example 1.} Let \( n = 6 \) and \( \sigma \in S_6 \) be defined by:
\[
\sigma(1) = 3, \; \sigma(3) = 1, \; \sigma(2) = 4, \; \sigma(4) = 2, \; \sigma(5) = 6, \; \sigma(6) = 5.
\]

\textbf{Step 1: Trace the orbits of \( \sigma \):}
\begin{itemize}
    \item Starting with \( 1 \): \( \sigma(1) = 3 \), \( \sigma(3) = 1 \). This forms the cycle \( (1 \, 3) \).
    \item Starting with \( 2 \): \( \sigma(2) = 4 \), \( \sigma(4) = 2 \). This forms the cycle \( (2 \, 4) \).
    \item Starting with \( 5 \): \( \sigma(5) = 6 \), \( \sigma(6) = 5 \). This forms the cycle \( (5 \, 6) \).
\end{itemize}

\textbf{Step 2: Write the cycle decomposition:}
\[
\sigma = (1 \, 3)(2 \, 4)(5 \, 6).
\]

\textbf{Example 2.} Let \( n = 5 \) and \( \sigma \in S_5 \) be defined by:
\[
\sigma(1) = 2, \; \sigma(2) = 3, \; \sigma(3) = 4, \; \sigma(4) = 5, \; \sigma(5) = 1.
\]

\textbf{Step 1: Trace the orbits of \( \sigma \):}
\begin{itemize}
    \item Starting with \( 1 \): \( \sigma(1) = 2 \), \( \sigma(2) = 3 \), \( \sigma(3) = 4 \), \( \sigma(4) = 5 \), \( \sigma(5) = 1 \). This forms the cycle \( (1 \, 2 \, 3 \, 4 \, 5) \).
\end{itemize}

\textbf{Step 2: Write the cycle decomposition:}
\[
\sigma = (1 \, 2 \, 3 \, 4 \, 5).
\]

\textbf{Counterexample.} Let \( \sigma \) be a mapping on \( \{1, 2, 3\} \) defined by \( \sigma(1) = 2 \), \( \sigma(2) = 2 \), \( \sigma(3) = 3 \).

\textbf{Violation of conditions:}
\begin{itemize}
    \item \( \sigma \) is not a bijection because \( \sigma(2) = 2 \) and no element maps to \( 1 \).
    \item Since \( \sigma \notin S_3 \), the cycle decomposition theorem does not apply.
\end{itemize}

\subsection{Chain of Theorems Learning Data}

\paragraph{Case 1: Angle Bisector Triangle Angle Relation}

\textbf{Question.} Given the theorem \textit{Angle Bisector Triangle Angle Relation}, analyze its definition, conditions for application, and key intuition.

\textbf{Source Theorems.}
\begin{enumerate}
    \item \textbf{Triangle Angle Sum Theorem:} The sum of the interior angles of a triangle is always \(180^\circ\).
    \item \textbf{Angle Bisector Definition:} If a segment bisects an angle of a triangle, it divides the triangle into two sub-triangles such that the bisected angle is split equally.
    \item \textbf{Exterior Angle Theorem:} The measure of an exterior angle of a triangle is equal to the sum of the measures of the two non-adjacent interior angles.
\end{enumerate}

\textbf{Theorem Composition.}
\begin{itemize}
    \item \textbf{Step 1: Apply the Triangle Angle Sum Theorem in sub-triangle \(ABD\).} Let triangle \(ABC\) have point \(D\) on side \(BC\), such that \(BD\) bisects \(\angle ABC\). In triangle \(ABD\), the sum of the angles is:
    \[
    \angle BAD + \angle DBA + \angle ADB = 180^\circ
    \]
    Rearranging to solve for \(\angle BAD\):
    \[
    \angle BAD = 180^\circ - \angle DBA - \angle ADB
    \]

    \item \textbf{Step 2: Apply the Angle Bisector Definition.} Since \(BD\) is the angle bisector of \(\angle ABC\), we know:
    \[
    \angle BAC = 2 \cdot \angle BAD
    \]

    \item \textbf{Step 3: Apply the Triangle Angle Sum Theorem in triangle \(ABC\).} In triangle \(ABC\), the sum of the angles is:
    \[
    \angle BAC + \angle ABC + \angle C = 180^\circ
    \]
    Substitute \(\angle BAC = 2 \cdot \angle BAD\) and solve for \(\angle C\):
    \[
    \angle C = 180^\circ - 2 \cdot \angle BAD - \angle ABC
    \]

    \item \textbf{Step 4: Eliminate intermediate variables using the relations derived.} Substitute \(\angle BAD = 180^\circ - \angle DBA - \angle ADB\) into the expression for \(\angle C\):
    \[
    \angle C = 180^\circ - 2 \cdot (180^\circ - \angle DBA - \angle ADB) - \angle ABC
    \]
    Simplify:
    \[
    \angle C = 180^\circ - 360^\circ + 2 \cdot (\angle DBA + \angle ADB) - \angle ABC
    \]
    \[
    \angle C = -180^\circ + 2 \cdot (\angle DBA + \angle ADB) - \angle ABC
    \]
\end{itemize}

\textbf{Definition.} Let triangle \(ABC\) have point \(D\) on side \(BC\), such that \(BD\) bisects \(\angle ABC\). If \(\angle DBA\) and \(\angle ADB\) are given, the measure of \(\angle C\) can be computed using:
\[
\angle C = -180^\circ + 2 \cdot (\angle DBA + \angle ADB) - \angle ABC
\]

\textbf{Conditions.}
\begin{itemize}
    \item \(BD\) must bisect \(\angle ABC\).
    \item Triangle \(ABC\) must be valid (sum of angles equals \(180^\circ\)).
    \item Angles \(\angle DBA\) and \(\angle ADB\) must be known.
\end{itemize}

\textbf{Functional Form.}
\[
\angle C = f(\angle DBA, \angle ADB, \angle ABC) = -180^\circ + 2 \cdot (\angle DBA + \angle ADB) - \angle ABC
\]

\textbf{Intuition (Formal Version).}
\begin{enumerate}
    \item \textbf{Define Objects:}
    \begin{itemize}
        \item Let triangle \(ABC\) have vertices \(A\), \(B\), and \(C\).
        \item Let point \(D\) lie on side \(BC\), such that \(BD\) bisects \(\angle ABC\).
        \item Let \(\angle DBA\) and \(\angle ADB\) be the given angles in sub-triangle \(ABD\).
    \end{itemize}

    \item \textbf{Apply Derived Theorem:}
    \begin{itemize}
        \item Use the Triangle Angle Sum Theorem in triangle \(ABD\) to compute \(\angle BAD\).
        \item Use the Angle Bisector Definition to compute \(\angle BAC = 2 \cdot \angle BAD\).
        \item Use the Triangle Angle Sum Theorem in triangle \(ABC\) to compute \(\angle C\).
    \end{itemize}

    \item \textbf{Compute:}
    \begin{itemize}
        \item Substitute all known values into the derived formula for \(\angle C\).
    \end{itemize}
\end{enumerate}

\textbf{Counterexample.}
\begin{itemize}
    \item \textbf{Failure Case:} If \(BD\) does not bisect \(\angle ABC\), the derived theorem fails because the assumption of equal division of \(\angle ABC\) into \(\angle BAD\) and \(\angle CAD\) is violated. For example, if \(BD\) is not an angle bisector but instead an arbitrary segment, the formula for \(\angle C\) will not hold.

    \item \textbf{Violated Condition:} Condition 1: \(BD\) must bisect \(\angle ABC\).
\end{itemize}

\paragraph{Case 2: Inscribed Angle and Triangle Angle Sum Relation in a Right Triangle}

\textbf{Question.} Given the theorem \textit{Inscribed Angle and Triangle Angle Sum Relation in a Right Triangle}, analyze its definition, conditions for application, and key intuition.

\textbf{Source Theorems.}
\begin{enumerate}
    \item \textbf{Triangle Angle Sum Theorem:} The sum of the interior angles of a triangle is \(180^\circ\).
    \item \textbf{Definition of a Right Triangle:} A right triangle contains one angle equal to \(90^\circ\).
    \item \textbf{Inscribed Angle Theorem:} An inscribed angle in a circle is half the measure of the corresponding central angle subtending the same arc.
\end{enumerate}

\textbf{Theorem Composition.}
\begin{itemize}
    \item \textbf{Step 1: Apply the Triangle Angle Sum Theorem.} Let \( \triangle ABC \) be a triangle where \( \angle ACB = 90^\circ \). By the Triangle Angle Sum Theorem:
    \[
    \angle A + \angle B + \angle C = 180^\circ
    \]
    Substituting \( \angle C = 90^\circ \), we get:
    \[
    \angle A + \angle B = 90^\circ
    \]

    \item \textbf{Step 2: Apply the Inscribed Angle Theorem.} Let point \( D \) lie on segment \( AB \), and let an arc be drawn with center \( B \) and radius \( BC \). The arc intersects \( AB \) at \( D \), forming an inscribed angle \( \angle ACD \). By the Inscribed Angle Theorem, \( \angle ACD \) is half the measure of the central angle \( \angle ABC \) subtending the same arc:
    \[
    \angle ACD = \frac{1}{2} \angle ABC
    \]

    \item \textbf{Step 3: Relate \( \angle ABC \) to \( \angle ACD \).} Rearranging the equation from Step 2, we find:
    \[
    \angle ABC = 2 \angle ACD
    \]

    \item \textbf{Step 4: Combine with the Triangle Angle Sum Relation.} From Step 1, \( \angle A + \angle B = 90^\circ \). Substituting \( \angle B = \angle ABC \), we get:
    \[
    \angle A + \angle ABC = 90^\circ
    \]
    Substituting \( \angle ABC = 2 \angle ACD \) (from Step 3), we find:
    \[
    \angle A + 2 \angle ACD = 90^\circ
    \]

    \item \textbf{Step 5: Eliminate Intermediate Variables.} Rearrange the equation to express \( \angle A \) in terms of \( \angle ACD \):
    \[
    \angle A = 90^\circ - 2 \angle ACD
    \]
\end{itemize}

\textbf{Definition.} Let \( \triangle ABC \) be a right triangle with \( \angle ACB = 90^\circ \). Let an arc be drawn with center \( B \) and radius \( BC \), intersecting segment \( AB \) at point \( D \). If \( \angle ACD \) is the inscribed angle formed by the arc, then the measure of \( \angle A \) is given by:
\[
\angle A = 90^\circ - 2 \angle ACD
\]

\textbf{Conditions.}
\begin{itemize}
    \item \( \triangle ABC \) is a right triangle with \( \angle ACB = 90^\circ \).
    \item Point \( D \) lies on segment \( AB \), and an arc is drawn with center \( B \) and radius \( BC \).
    \item \( \angle ACD \) is the inscribed angle formed by the arc.
\end{itemize}

\textbf{Functional Form.}
\[
\angle A = 90^\circ - 2 \angle ACD
\]

\textbf{Intuition (Formal Version).}
\begin{enumerate}
    \item Let \( \triangle ABC \) be a right triangle with \( \angle ACB = 90^\circ \).
    \item Let an arc be drawn with center \( B \) and radius \( BC \), intersecting segment \( AB \) at point \( D \).
    \item By the Inscribed Angle Theorem, \( \angle ACD \) is half the measure of the central angle \( \angle ABC \): \( \angle ACD = \frac{1}{2} \angle ABC \).
    \item Using the Triangle Angle Sum Theorem, \( \angle A + \angle ABC = 90^\circ \).
    \item Substituting \( \angle ABC = 2 \angle ACD \), we find \( \angle A = 90^\circ - 2 \angle ACD \).
\end{enumerate}

\textbf{Example 1.}
\begin{itemize}
    \item \textbf{Step 1: Define Objects.} Let \( \triangle ABC \) be a right triangle with \( \angle ACB = 90^\circ \). Let \( \angle ACD = 20^\circ \).

    \item \textbf{Step 2: Apply Derived Theorem.} Using the derived theorem:
    \[
    \angle A = 90^\circ - 2 \angle ACD
    \]
    Substituting \( \angle ACD = 20^\circ \):
    \[
    \angle A = 90^\circ - 2(20^\circ) = 50^\circ
    \]

    \item \textbf{Step 3: Compute.} The measure of \( \angle A \) is \( 50^\circ \).
\end{itemize}

\textbf{Counterexample.}
\begin{itemize}
    \item \textbf{Failure Case:} If \( \triangle ABC \) is not a right triangle (i.e., \( \angle ACB \neq 90^\circ \)), the derived theorem does not hold because the relationship between \( \angle A \) and \( \angle ACD \) depends on the right angle at \( C \). For example, let \( \angle ACB = 80^\circ \) and \( \angle ACD = 20^\circ \). The derived theorem incorrectly predicts \( \angle A = 90^\circ - 2(20^\circ) = 50^\circ \), but the actual value of \( \angle A \) depends on the specific geometry of the triangle and cannot be determined using this theorem.

    \item \textbf{Violated Condition:} Condition 1: \( \triangle ABC \) must be a right triangle with \( \angle ACB = 90^\circ \).
\end{itemize}

%% file: checklist.tex
\section*{NeurIPS Paper Checklist}

\begin{enumerate}

\item {\bf Claims}
    \item[] Question: Do the main claims made in the abstract and introduction accurately reflect the paper's contributions and scope?
    \item[] Answer:  \answerYes{}
    \item[] Justification: The abstract and introduction clearly describe the motivation, proposed Theorem-SFT framework, and its empirical improvements over vanilla SFT on mathematical reasoning benchmarks. The scope and limitations of the method are also discussed in the paper.

\item {\bf Limitations}
    \item[] Question: Does the paper discuss the limitations of the work performed by the authors?
    \item[] Answer: \answerYes{}
    \item[] Justification: We have limitations~\ref{sec:limitation}

\item {\bf Theory assumptions and proofs}
    \item[] Question: For each theoretical result, does the paper provide the full set of assumptions and a complete (and correct) proof?
    \item[] Answer:  \answerNA{}
    \item[] Justification: The paper does not contain formal theoretical results or mathematical proofs. Our contributions focus on framework design, data construction, and empirical evaluation.

    \item {\bf Experimental result reproducibility}
    \item[] Question: Does the paper fully disclose all the information needed to reproduce the main experimental results of the paper to the extent that it affects the main claims and/or conclusions of the paper (regardless of whether the code and data are provided or not)?
    \item[] Answer:\answerYes{}
    \item[] Justification: Sections~\ref{sec:theorem-sft} and~\ref{sec:exe} describe the method and experimental setup, and the appendix provides data-generation prompts and examples. Code, data, and trained model checkpoints are planned for release at \url{https://github.com/Ivine11/Theorem-SFT.git}.

\item {\bf Open access to data and code}
    \item[] Question: Does the paper provide open access to the data and code, with sufficient instructions to faithfully reproduce the main experimental results, as described in supplemental material?
    \item[] Answer:\answerYes{}
    \item[] Justification: The release repository is \url{https://github.com/Ivine11/Theorem-SFT.git}. Code, data, and trained model checkpoints will be made available there; the paper currently provides data-generation prompts and examples.

\item {\bf Experimental setting/details}
    \item[] Question: Does the paper specify all the training and test details (e.g., data splits, hyperparameters, how they were chosen, type of optimizer) necessary to understand the results?
    \item[] Answer: {\answerNo{}} 
    \item[] Justification: Due to space limitations, we provide the main experimental pipeline and key hyperparameters in Sec.~\ref{sec:exe} and \ref{sec:theo_learn}. These include the training setup, optimization strategy, and evaluation protocol, which are sufficient to understand the reported results.

\item {\bf Experiment statistical significance}
    \item[] Question: Does the paper report error bars suitably and correctly defined or other appropriate information about the statistical significance of the experiments?
    \item[] Answer: \answerNo{} 
    \item[] Justification: We report single-run accuracy results across multiple datasets and models under consistent evaluation settings to compare performance. In addition, we analyze the output probability distributions of Vanilla SFT and baselines to provide qualitative insights into the observed behavior. However, we do not include multiple-run statistics or error bars due to computational constraints.Please check in Sec.~\ref{sec:whyFail} and ~\ref{sec:exe}

\item {\bf Experiments compute resources}
    \item[] Question: For each experiment, does the paper provide sufficient information on the computer resources (type of compute workers, memory, time of execution) needed to reproduce the experiments?
    \item[] Answer: \answerYes{} 
    \item[] Justification: Appendix~\ref{sec:compute} reports the use of eight NVIDIA H100 GPUs and eight NVIDIA H20 GPUs and describes the training setup. Per-experiment execution times and total GPU-hours are not reported.
    
\item {\bf Code of ethics}
    \item[] Question: Does the research conducted in the paper conform, in every respect, with the NeurIPS Code of Ethics \url{https://neurips.cc/public/EthicsGuidelines}?
    \item[] Answer:\answerYes{} 
    \item[] Justification: The research in this paper uses publicly available datasets and standard machine learning training procedures. It does not involve human subjects, personal data, or sensitive information, and therefore complies with the NeurIPS Code of Ethics.

\item {\bf Broader impacts}
    \item[] Question: Does the paper discuss both potential positive societal impacts and negative societal impacts of the work performed?
    \item[] Answer:\answerNo{}
    \item[] Justification: The paper focuses on mathematical reasoning experiments and does not include a discussion of both positive and negative societal impacts.
    
\item {\bf Safeguards}
    \item[] Question: Does the paper describe safeguards that have been put in place for responsible release of data or models that have a high risk for misuse (e.g., pre-trained language models, image generators, or scraped datasets)?
    \item[] Answer:\answerNA{} 
    \item[] Justification: This work does not involve the release of new large-scale generative models or datasets with high risk of misuse. The experiments are conducted on publicly available benchmarks and standard language model fine-tuning setups, and therefore no additional safeguards are required.

\item {\bf Licenses for existing assets}
    \item[] Question: Are the creators or original owners of assets (e.g., code, data, models), used in the paper, properly credited and are the license and terms of use explicitly mentioned and properly respected?
    \item[] Answer: \answerYes{}
    \item[] Justification: We use publicly available datasets and pre-trained models, which are properly cited in the paper. All assets used in this work follow their original licenses and terms of use, and we ensure appropriate attribution to the original creators.

\item {\bf New assets}
    \item[] Question: Are new assets introduced in the paper well documented and is the documentation provided alongside the assets?
    \item[] Answer: \answerYes{}
    \item[] Justification: The paper documents theorem-data generation through descriptions, prompts, and examples. The repository at \url{https://github.com/Ivine11/Theorem-SFT.git} is designated for the planned release of code, data, and trained model checkpoints with accompanying documentation.

\item {\bf Crowdsourcing and research with human subjects}
    \item[] Question: For crowdsourcing experiments and research with human subjects, does the paper include the full text of instructions given to participants and screenshots, if applicable, as well as details about compensation (if any)? 
    \item[] Answer: \answerNA{}
    \item[] Justification: This work does not involve crowdsourcing experiments or human subject studies, and therefore no participant instructions, screenshots, or compensation details are applicable.

\item {\bf Institutional review board (IRB) approvals or equivalent for research with human subjects}
    \item[] Question: Does the paper describe potential risks incurred by study participants, whether such risks were disclosed to the subjects, and whether Institutional Review Board (IRB) approvals (or an equivalent approval/review based on the requirements of your country or institution) were obtained?
    \item[] Answer: \answerNA{} 
    \item[] Justification: This work does not involve human subjects, user studies, or data collection from individuals. Therefore, Institutional Review Board (IRB) approval is not required.

\item {\bf Declaration of LLM usage}
    \item[] Question: Does the paper describe the usage of LLMs if it is an important, original, or non-standard component of the core methods in this research? Note that if the LLM is used only for writing, editing, or formatting purposes and does \emph{not} impact the core methodology, scientific rigor, or originality of the research, declaration is not required.
    \item[] Answer:\answerYes{}
    \item[] Justification: GPT-5 was used to generate theorem-learning materials and abstract theorem chains and to check generated materials for mathematical errors. Model-based checks do not constitute formal verification. The paper also identifies the language models used for fine-tuning and evaluation.

\end{enumerate}